\documentclass{new_tlp}
\usepackage{mathptmx}

\usepackage{soul}
\usepackage{url}
\usepackage[hidelinks]{hyperref}
\usepackage[utf8]{inputenc}
\usepackage[small]{caption}
\usepackage{graphicx}
\usepackage{amsmath}
\usepackage{booktabs}
\usepackage{algorithm}
\usepackage{algorithmic}
\usepackage{xcolor}

\usepackage{listings}
\lstdefinelanguage{asp}
{morekeywords={in,out,assumption,supported,head,body,contrary,defeated,triggered_by_in,derivable_from_undefeated,triggered_by_undefeated,attacked_by_undefeated,sentence,derivable,applicable_rule,triggered_rule,rule,preference,optimize,preferred,less_preferred,no_less_preferred,pref_supported,pref_triggered_by_in,normally_defeated,derived_from_undefeated_assumption,triggered_by_undefeated_assumption,in_attacked_by_norm_undef,reversely_defeated,suspect,other,target,preferredly_sup_by_suspects,preferredly_trig_suspects,target_normally_attacked,derivable_from_target,triggered_by_target,suspect_attacked_by_target,target_reversely_attacked,supported_by_def,def,triggered_by_def,suspect_normally_defeated_by_def,supported_by_suspects,trig_suspects,reaches_via_suspect,suspect_reversely_defeated_by_def,suspect_attacked_by_target,iteration,n_assumptions,count,in_attacked_by_normally_undefeated,undefeated,pref_supported_by_suspects,pref_triggered_by_suspects,in_normally_defeated_by_suspects,supported_by_in,reach_in,in_reversely_defeated_by_suspects,suspect_normally_defeated_by_in,suspect_reversely_defeated_by_in,suspect_attacked_by_target,supported_by_in,triggered_by_suspects,reach_suspect,pref_supported_by_undefeated,pref_triggered_by_undefeated,undefeated_normally_defeated_by_undefeated,undefeated_reversely_defeated_by_undefeated,pref_triggered_by_suspects,undef_asm_att_by_norm_undef_asm,derived_from_undefeated}, 
literate={:-}{{$\la\ $}}1 {not}{{$\naf$}}1 {\\el}{{$\in$}}1 {\\cup}{{$\cup$}}1 {:~}{{$\law$}}1,
morecomment=[l]{\%},
}

\newtheorem{theorem}{Theorem}
\newtheorem{proposition}[theorem]{Proposition}

\newtheorem{definition}{Definition}
\newtheorem{example}{Example}

\newcommand{\abaplus}{ABA$^+$}

\newcommand{\contrary}[1]{\overline{#1}}
\newcommand{\contraryempty}{\contrary{\phantom{a}}}

\newcommand{\thc}{\mathit{Th}}
\newcommand{\cf}{\mathit{cf}}

\newcommand{\adm}{\mathit{adm}}
\newcommand{\pref}{\mathit{prf}}

\newcommand{\comp}{\mathit{com}}

\newcommand{\aspassump}{{\bf assumption}}
\newcommand{\asphead}{{\bf head}}
\newcommand{\aspbody}{{\bf body}}
\newcommand{\aspcontrary}{{\bf contrary}}
\newcommand{\aspin}{{\bf in}}
\newcommand{\aspout}{{\bf out}}
\newcommand{\aspsupported}{{\bf supported}}

\newcommand{\asppreferred}{{\bf preferred}}

\newcommand{\aspmodule}[1]{\pi_{\mathit{#1}}}
\newcommand{\naf}{{\it not}\,}
\newcommand{\la}{\leftarrow}

\begin{document}

\title{Harnessing Incremental Answer Set Solving for\\ Reasoning in Assumption-Based Argumentation\thanks{Work financially supported by Academy of Finland
(grant 322869), University of Helsinki Doctoral Programme in Computer Science DoCS,
 and the Austrian Science Fund (FWF): P30168-N31 and I2854.}}
\shorttitle{Incremental ASP Solving for ABA}

\author[T. Lehtonen, J. P. Wallner and M. J\"arvisalo]{TUOMO LEHTONEN\\University of Helsinki, Finland\\\email{tuomo.lehtonen@helsinki.fi} 
\and
JOHANNES P. WALLNER\\Graz University of Technology, Austria\\\email{wallner@ist.tugraz.at}
\and
MATTI J\"ARVISALO\\University of Helsinki, Finland\\\email{matti.jarvisalo@helsinki.fi}}

\maketitle

\begin{abstract}
Assumption-based argumentation (ABA) is a central structured argumentation formalism.
As shown recently, answer set programming (ASP) enables
efficiently solving NP-hard reasoning tasks of ABA in practice, in particular
in the commonly studied logic programming fragment of ABA.
In this work, we harness recent advances in incremental ASP solving
for developing  effective algorithms for reasoning
tasks in the logic programming fragment of ABA that are presumably hard for the second  level of the polynomial hierarchy,
including skeptical reasoning under preferred semantics as well as preferential reasoning.
In particular, we develop non-trivial counterexample-guided abstraction refinement procedures
based on incremental ASP solving for these tasks. We also show empirically that
the procedures are significantly more effective than previously proposed algorithms for the tasks.

\emph{This paper is under consideration for acceptance in TPLP.}
\end{abstract}


\section{Introduction}

Argumentation, and in particular the study of computational models of argument, 
constitutes a core research area in artificial intelligence, and  knowledge representation
and non-monotonic reasoning in particular~\cite{arguHandbook}.
Computational models for structured argumentation, as opposed to abstract argumentation,
make the internal structure of arguments explicit, supporting the view
that arguments are most often made explicit through derivations from more
basic structures, thereby having an intrinsic structure. 
Various structured argumentation formalisms have been proposed, each with their own
features and applications~\cite{CyrasFST2018,ModgilP18,BesnardH18,GarciaS18}. 

In this work we focus on the commonly studied logic programming fragment of 
assumption-based argumentation (ABA)~\cite{CyrasFST2018} 
and its extension, ABA$^+$, equipped with preferences~\cite{CyrasT16}.
These central structured argumentation formalisms have found applications e.g., in decision making in a multi-agent context~\cite{DBLP:conf/atal/FanTMW14}, game
theory~\cite{Fan:2016:IGA:2936924.2936964}, and in choosing treatment recommendations based
 on clinical guidelines and preferential information given by patients~\cite{CyrasO19}.
In addition to applications, the challenge of developing
efficient systems for ABA reasoning is also highlighted by the 2021 ICCMA argumentation
system competition, which for the first time called for ABA reasoning systems for several NP-hard problems.

Among the algorithmic approaches proposed for reasoning in ABA~\cite{DungKT06,GartnerT2007b,Toni13,CravenTW13,CravenT16,LehtonenWJ17,LehtonenWJ:JAIR2021},
in terms of scalability arguably the currently most efficient practical approach
is based on encoding ABA reasoning tasks declaratively using answer set programming
(ASP)~\cite{DBLP:conf/iclp/GelfondL88,DBLP:journals/amai/Niemela99}, and invoking off-the-shelf ASP solvers for the reasoning part~\cite{LehtonenWJ:JAIR2021,CaminadaS17}.
While this approach is noticeably more efficient than other competing 
ABA reasoning systems  on NP-complete variants of ABA reasoning, the approach
was not directly extended to cover all \emph{beyond-NP} variants of ABA reasoning, i.e.,
reasoning tasks which are presumably hard for the second-level of the polynomial 
hierarchy. In particular, skeptical acceptance in ABA under preferred semantics was treated resorting
to the so-called Asprin approach, although a direct treatment would be viable. Further, 
credulous reasoning in ABA$^+$ under admissible and complete semantics was not covered.

Motivated by the success of ASP-based ABA reasoning, in this work
we harness very recent advances in incremental ASP solving~\cite{DBLP:journals/corr/abs-2008-06692,DBLP:journals/aicom/GebserKKOSS11} 
for developing counterexample-guided abstraction refinement~\cite{DBLP:journals/jacm/ClarkeGJLV03,DBLP:journals/tcad/ClarkeGS04} style algorithms
for skeptical reasoning in ABA under
preferred semantics, as well as credulous reasoning in ABA$^+$
 under admissible and complete semantics. Compared to the
 currently existing ABA reasoning systems supporting these tasks, 
in particular the Asprin-based~\cite{BrewkaD0S15} approach to reasoning in ABA under preferred semantics~\cite{LehtonenWJ:JAIR2021},
and the ABAplus system~\cite{BaoCT17} for enumerating admissible and complete assumption sets in ABA$^+$, 
our approach provides significant performance improvements in practice and allows
for directly reasoning about credulous acceptance in ABA$^+$.
Our implementation is available at \url{https://bitbucket.org/coreo-group/aspforaba}.

\section{Assumption-Based Argumentation}

\label{sec:aba}

We recall assumption-based argumentation (ABA)~\cite{BondarenkoDKT97,Toni14,CyrasFST2018} and \abaplus{}~\cite{CyrasT16,CyrasT16-nmr,BaoCT17,DBLP:phd/ethos/Cyras17} which extends ABA with preferences over assumptions. 
We define \abaplus{} frameworks, as \abaplus{} is a generalization of ABA.

We focus on the commonly studied logic programming fragment of ABA and \abaplus{}. In particular,
we assume a deductive system $(\mathcal{L},\mathcal{R})$ with $\mathcal{L}$ a set of atoms and $\mathcal{R}$ a set of inference rules over $\mathcal{L}$ with a rule $r \in \mathcal{R}$ having the form
$a_0 \leftarrow a_1,\ldots,a_n$ with $a_i \in \mathcal{L}$. 
To distinguish ABA atoms from atoms in answer set programming, we refer to the former as sentences. We denote the head of rule $r$ by $head(r) = \{a_0\}$ and the (possibly empty) body of $r$ by $body(r) = \{a_1,\ldots,a_n\}$. 

An \abaplus{} framework is a tuple $F=(\mathcal{L},\mathcal{R},\mathcal{A},\contraryempty,\leq)$ with $(\mathcal{L},\mathcal{R})$ a deductive system, 
a set of assumptions $\mathcal{A} \subseteq \mathcal{L}$, a function $\contraryempty$ mapping assumptions $\mathcal{A}$ to sentences $\mathcal{L}$, and a 
preorder $\leq$ on $\mathcal{A}$. 
The strict counterpart $<$ of $\leq$ is defined as usual by
$a < b$ iff $a \leq b$ and $b \not \leq a$, for $a,b \in \mathcal{A}$.
An ABA framework, that is \abaplus{} without preferences, is an \abaplus{} framework with $\leq\ = \emptyset$, denoted by $(\mathcal{L},\mathcal{R},\mathcal{A},\contraryempty)$.
In this paper, we focus on so-called flat \abaplus{} frameworks where assumptions cannot be derived, i.e., do not occur in heads of rules. 
We assume that each set in an \abaplus{} is finite.

There are two notions of derivations for ABA: tree-derivations and forward-derivations.
\abaplus{} is in general  defined through tree-derivations.
We briefly recall both notions. A sentence $s \in \mathcal{L}$ is tree-derivable from a set of assumptions $X \subseteq \mathcal{A}$
and rules $R \subseteq \mathcal{R}$, denoted by $X \models_R s$, if there is a finite tree with the root labeled by $s$, the leaves
labeled by elements of $X$, and for each internal node there is a rule $r \in R$ such that the node itself is labeled by $head(r)$ and the set of labels of the children of this node is $body(r)$. For each rule $r \in R$ there is a node labeled in this way.
For brevity, $R$ can be left unspecified and be assumed to be some suitable subset of $\mathcal{R}$.
A sentence $a \in \mathcal{L}$ is forward-derivable from a set $X \subseteq \mathcal{A}$ via rules $\mathcal{R}$,
denoted by $X \vdash_{\mathcal{R}} a$, if there is a sequence of rules $(r_1,\ldots,r_n)$ such that $head(r_n) = a$, for
each rule $r_i$ we have $r_i \in \mathcal{R}$, and each sentence in the body of $r_i$ is derived from rules earlier in the
sequence or is in $X$, i.e., $body(r_i) \subseteq X \cup \bigcup_{j < i}head(r_j)$.
The deductive closure for an assumption set $X$ w.r.t.\ rules $\mathcal{R}$ is given by $\thc_{\mathcal{R}}(X) = \{a \mid X \vdash_{\mathcal{R}} a\}$. 

\begin{definition}
Let $(\mathcal{L},\mathcal{R},\mathcal{A},\contraryempty,\leq)$ be an \abaplus{} framework, and
$A,B \subseteq \mathcal{A}$ be two sets of assumptions. 
$A$ $<$-attacks $B$ if
\begin{itemize}
 \item $A' \models_R \contrary{b}$ for some $A' \subseteq A$, $b\in B$, and $\not \exists a' \in A'$ with $a' < b$, or
 \item $B' \models_R \contrary{a}$ for some $a \in A$ and $B' \subseteq B$ s.t.\ $\exists b' \in B'$ with $b' < a$.
\end{itemize}
\end{definition}
In words, set $A$ attacks $B$ if (i)~from a subset $A'$ of $A$, one can tree-derive a contrary of an assumption $b\in B$ and no member in $A'$ is strictly less preferred than $b$, or (ii)~from $B$, via subset $B'$ one can tree-derive a contrary of an assumption $a\in A$ and some member of $B'$ is strictly less preferred than $a$. Attacks of type (i) are  \emph{normal} $<$-attacks and those of type (ii) \emph{reverse} $<$-attacks, with the intuition that the (non-preference based) conflict in (i) succeeds and in case of (ii) is countered and reversed  by the preference relation. 
For brevity, we omit set notation when $A$ $<$-attacks a singleton $\{b\}$ (then we say $A$ $<$-attacks $b$).

\begin{definition}
Let $F=(\mathcal{L},\mathcal{R},\mathcal{A},\contraryempty,\leq)$ be an \abaplus{} framework. 
An assumption set $A\subseteq \mathcal{A}$ is called conflict-free if $A$ does not $<$-attack itself. 
Set $A$ defends assumption set $B \subseteq \mathcal{A}$ if for all $C \subseteq \mathcal{A}$ that $<$-attack $B$ it holds that $A$ $<$-attacks $C$. 
\end{definition}

\begin{definition}
Let $F=(\mathcal{L},\mathcal{R},\mathcal{A},\contraryempty,\leq)$ be an \abaplus{} framework. 
Further, let $A \subseteq \mathcal{A}$ be a conflict-free set of assumptions in $F$. Set $A$ is 
\begin{itemize}
 \item $<$-\emph{admissible} in $F$ if $A$ defends itself; 
 \item $<$-\emph{complete} in $F$ if $A$ is admissible in $F$ and contains every assumption set defended by A; 
 \item $<$-\emph{preferred} in $F$ if $A$ is $<$-admissible and there is no $<$-admissible set of assumptions $B$ in $F$ with $A \subset B$. 
\end{itemize}
\end{definition}

We use the term \emph{$<$-$\sigma$ assumption set} for an assumption set under  a semantics $\sigma \in \{\adm$, $\comp$, $\pref\}$, i.e., $<$-admissible, $<$-complete, and $<$-preferred assumption set, respectively. 

For ABA, we refer to the corresponding semantics without the preference relation $<$ (e.g., complete semantics instead of $<$-complete semantics). 
Attacks ($\emptyset$-attacks) in ABA frameworks simplify to attacks from $A$ to $B$ when $A \vdash_{\mathcal{R}} \overline{b}$ for $b \in B$. 

Main reasoning tasks on \abaplus{} are the following. 
\begin{definition}
Let $F=(\mathcal{L},\mathcal{R},\mathcal{A},\contraryempty,\leq)$ be an \abaplus{} framework and $<$-$\sigma$ a semantics. 
A sentence $s \in \mathcal{L}$ is 
\begin{itemize}
 \item credulously accepted in $F$ under $<$-$\sigma$ if 
there is a $<$-$\sigma$ assumption set $A$ s.t.\ $s \in \thc_{\mathcal{R}}(A)$; and 
 \item skeptically accepted in $F$ under $<$-$\sigma$ if 
  $s \in \thc_{\mathcal{R}}(A)$ for all $<$-$\sigma$ assumption sets $A$. 
\end{itemize}
\end{definition}
The tasks for ABA are analogous (disregarding $<$). 

\begin{example}
Let $F$ be an \abaplus{} framework with $\mathcal{A} = \{a,b,c\}$, $\overline{b}=x$, $\overline{c}=y$, $a<c$, and $\mathcal{R} = \{(x \leftarrow a), (y \leftarrow a) \}$. 
We have $\{a\}$ normally $<$-attacks $b$ and $\{c\}$ reversely $<$-attacks $a$.
The $<$-admissible sets of $F$ are $\emptyset$, $\{c\}$, and $\{b,c\}$, and the framework has the unique $<$-complete set $\{b,c\}$.
\end{example}

We focus on computationally hard reasoning tasks in ABA and \abaplus{}. 
Deciding skeptical acceptance under preferred semantics in ABA is $\Pi^P_2$-complete~\cite{DBLP:journals/ai/DimopoulosNT02}. 
In \abaplus{}, 
credulous acceptance under $<$-admissible semantics is $\Sigma^P_2$-complete, and checking whether a set is $<$-admissible and $<$-complete is coNP-complete and coNP-hard, respectively~\cite{LehtonenWJ:JAIR2021}.

\section{Algorithms}

We present ASP-based counterexample-guided abstraction refinement (CEGAR) algorithms for  beyond-NP reasoning tasks in ABA and \abaplus{}. 
The CEGAR-based algorithms follow the iterative schema of considering an NP-abstraction of the solution space (containing spurious solutions), and drawing candidates from this space. 
At each iteration, a candidate solution is obtained (if one remains) by calling an ASP solver.
If no further candidate solutions can be obtained, the search terminates.
If a candidate is obtained, one checks with another ASP solver call whether the candidate
 is an actual solution. If it is, the search terminates. 
If not, a counterexample is obtained, and
the abstraction is refined (solution space is reduced) by ruling out from further
consideration at least the candidate solution. 

We briefly recap basic ASP concepts. An answer set program $\pi$ consists of rules $r$ of the form
$h \la b_1,\ldots,b_k,\naf b_{k+1},\ldots,\ \naf b_m$, where  $h$ and each $b_i$ is an atom. 
A literal is an atom or a default negated ($\naf$) atom. 
A rule is positive if $k=m$, a fact if $m=0$, and a constraint if there is no head $h$ (then a shorthand for the same rule with a fresh atom $a$ in the head and default negated in the body).
An atom $b_i$ has the form $p(t_1,\ldots,t_n)$ with $p$ a predicate and with each $t_j$  a constant or a variable. 
An answer set program, a rule, and an atom, respectively, is ground if it is free of variables. 
For a non-ground program,  $GP$ is the set of rules obtained by applying all possible
substitutions from the variables to the set of constants appearing in the program. 
An interpretation $I$, i.e., a subset of all the ground atoms, satisfies a positive rule $r = h \la b_1,\ldots,b_k$ 
iff all positive body elements $b_1,\ldots,b_k$ are in $I$ implies that the head atom is in $I$. 
For a program $\pi$ consisting only of positive rules, let $Cl(\pi)$ be the uniquely determined interpretation $I$ 
that satisfies all rules in $\pi$ and no subset of $I$ satisfies all rules in $\pi$. 
Interpretation $I$ is an answer set of a ground program $\pi$ if $I = Cl(\pi^I)$ where  $\pi^I = \{(h \la b_1,\ldots,b_k) \mid (h \la b_1,\ldots,b_k,\naf b_{k+1},\ldots, \naf b_m) \in \pi, \{b_{k+1},\ldots,b_m\} \cap I = \emptyset\}\}$ is the reduct; and 
of a non-ground program $\pi$ if $I$ is an answer set of $GP$ of $\pi$. A program $\pi$ is satisfiable iff there is an answer set of $\pi$. 

We make use of the following shorthands. Let $I$ be an interpretation (set of ASP atoms), ${\bf p}$ be some ASP predicate of arity one, and $M=\{l_1,\ldots,l_n\}$ a set of ASP literals. We define 
 ${\bf p}(I) = \{{\bf p}(x) \mid {\bf p}(x) \in I\}$, and 
 $\mathit{constr}(M) =\ \la l_1,\ldots,l_n$.
That is, ${\bf p}(I)$ is a set of atoms $ {\bf p}(x)$ which are contained in the interpretation, and $\mathit{constr}(M)$ is an ASP constraint containing $M$ as its body. 
For reasons of convenience, if $l$ is an ASP literal, we also define the shorthand $\mathit{constr}(l) =\ \la l$ (i.e., allowing $M$ to be a set or a single literal). 

In the following we refer to forward-derivations when talking about derivations.
For ABA, tree and forward-derivations are equivalent for the problems considered here~\cite{DungKT06,DungTM10}.
While forward-derivations are not directly applicable for \abaplus, for our approach to $<$-admissible and $<$-complete semantics we employ previous results~\cite{LehtonenWJ:JAIR2021} together with new ones (Section~\ref{sec:aba+prop}), which allow for avoiding naive application of tree-derivations which would require explicitly constructing arguments.

\begin{algorithm}[!b]
\caption{Skeptical acceptance under preferred}
\label{alg:aba-preferred-skept}
\begin{algorithmic}[1]
\REQUIRE ABA framework $F=(\mathcal{L},\mathcal{R},\mathcal{A},\contraryempty)$, $s \in \mathcal{L}$
    \ENSURE return YES if $s$ is skeptically accepted under preferred semantics in $F$, NO otherwise
    \STATE{$\pi := \mathtt{ABA}(F) \cup \aspmodule{com}$}
    \STATE{$\algorithmicwhile\ \pi \cup \{\mathit{constr}(\aspsupported(s))\}$ is satisfiable $\algorithmicdo$}\label{alg:prf-skept-2}
    \STATE{\hspace{\algorithmicindent} Let $I$ be the found answer set}
    \STATE{\hspace{\algorithmicindent} $\pi := \pi \cup \{\mathit{constr}(\aspout(I))\}$} 
    \STATE{\hspace{\algorithmicindent} $\algorithmicwhile\ \pi \cup \{\mathit{constr}(\aspsupported(s))\} \cup \aspin(I)$ is satisfiable$\ \algorithmicdo$} \label{alg:prf-skept-5} 
    \STATE{\hspace{\algorithmicindent}\hspace{\algorithmicindent} Let $I$ be the found answer set}
    \STATE{\hspace{\algorithmicindent}\hspace{\algorithmicindent} $\pi :=\pi \cup \{\mathit{constr}(\aspout(I))\}$} 
    \STATE{\hspace{\algorithmicindent} $\algorithmicif\ \pi\cup \aspin(I)$ is unsatisfiable $\algorithmicthen\ \algorithmicreturn$ NO} \label{alg:prf-skept-9}
    \RETURN YES
\end{algorithmic}
\end{algorithm}

\subsection{Skeptical Acceptance under Preferred Semantics} 

We begin with skeptical reasoning under preferred semantics in ABA, which is a $\Pi^P_2$-complete problem.
Following successful schemes for the same reasoning task on abstract argumentation frameworks (AFs)~\cite{CeruttiGTW2018}, we present Algorithm~\ref{alg:aba-preferred-skept} for deciding skeptical acceptance of sentences in an ABA framework.
We encode the given ABA framework $F=(\mathcal{L},\mathcal{R},\mathcal{A},\contraryempty)$ as ASP: assign each rule a unique identifier ($\mathcal{R}=\{r_1,...,r_n\}$) and let 
\begin{align*}
\mathtt{ABA}(F) =
&\{ \aspassump(a). \mid a \in \mathcal{A}  \}\ \cup\ \\
&\{ \asphead(i,b). \mid r_i \in \mathcal{R}, b \in head(r_i)  \}\ \cup\\
&\{ \aspbody(i,b). \mid r_i \in \mathcal{R}, b \in body(r_i)  \}\ \cup \\
&\{ \aspcontrary(a,x). \mid x = \overline{a}, a \in \mathcal{A}  \}.
\end{align*}
Listing~\ref{asp:com} presents module $\pi_{\comp}$ for finding complete assumption sets, including a possible queried sentence~\cite{LehtonenWJ:JAIR2021}. 
Now $I$ is an answer set of $\mathtt{ABA}(F) \cup \aspmodule{com}$ iff $\{ a \mid \aspin(a) \in I\}$ is a complete assumption set of $F$, and 
one can derive the sentence $x$ from this complete assumption set iff $\aspsupported(x) \in I$. Further, $\aspout(I)$ contains the assumptions of $F$ that are not part of $\aspin(I)$. 

\begin{lstlisting}[caption={Module $\aspmodule{\comp}$},frame=lines,label=asp:com,float]
in(X) :- assumption(X), not out(X).
out(X) :- assumption(X), not in(X).
supported(X) :- assumption(X), in(X).
supported(X) :- head(R,X), triggered_by_in(R).
triggered_by_in(R) :- head(R,_), supported(X) : body(R,X).
:- in(X), contrary(X,Y), supported(Y).
defeated(X) :- supported(Y), contrary(X,Y).
derived_from_undefeated(X) :- assumption(X), not defeated(X).
derived_from_undefeated(X) :- head(R,X), triggered_by_undefeated(R).
triggered_by_undefeated(R) :- head(R,_), derived_from_undefeated(X) : body(R,X).
attacked_by_undefeated(X) :- contrary(X,Y), derived_from_undefeated(Y).
:- in(X), attacked_by_undefeated(X).
:- out(X), not attacked_by_undefeated(X).
\end{lstlisting}

Algorithm~\ref{alg:aba-preferred-skept} decides skeptical acceptance under preferred semantics for ABA frameworks by first generating a complete assumption set\footnote{We remark that one can also use admissible sets instead of complete sets in this algorithm.} within the framework that does not derive the queried sentence $s$ (Line~\ref{alg:prf-skept-2}).  
If there is no answer set found in the first application of the while loop in Line~\ref{alg:prf-skept-2}, then all complete assumption sets of $F$ derive $s$, and the algorithm terminates. 
Otherwise, we add to the ASP encoding $\pi$ the constraint ruling out the complete set encoded in $\aspin(I)$ as a solution: we add $\mathit{constr}(\aspout(I))$, which states that at least one atom in $\aspout(I)$ must not be present in an answer set from now on (excluding $\aspin(I)$ and its subsets). 
Subsequently, we iteratively generate proper supersets of a currently found complete assumption set not deriving $s$ (loop starting in Line~\ref{alg:prf-skept-5}). 
When this inner loop terminates, we found a complete assumption set that is $\subseteq$-maximal among all complete assumption sets that do not derive $s$, and $\pi$ currently contains the last constraint ruling out this particular complete assumption set and its subsets. In Line~\ref{alg:prf-skept-9}, we check with an ASP solver call 
whether $\pi \cup \aspin(I) := \pi \cup \{\aspin(a). \mid \aspin(a) \in I\}$ is satisfiable. If it is, there is a complete assumption set that is a proper \emph{superset} of the assumptions encoded in $\aspin(I)$ and
that derives $s$. 
In this case $\aspin(I)$ is not a preferred assumption set and thus not a counterexample to $s$ being skeptically accepted under preferred semantics, so the algorithm proceeds to searching for a new candidate (Line~\ref{alg:prf-skept-2}).
Importantly,  $\pi$ still contains the constraints ruling out any subset of $\aspin(I)$.
On the other hand, if in Line~\ref{alg:prf-skept-9} the ASP solver reports unsatisfiability,  $\aspin(I)$ represents a preferred assumption set not deriving $s$, constituting a counterexample to $s$ being skeptically accepted under preferred semantics. 
To enumerate all preferred assumption sets, it suffices to omit the query and Line 8, and collect each answer after exiting the inner loop; see~\ref{appendix:algs}.

The following proposition states the correctness of the approach; correctness follows by the previous discussion on the details of the algorithm and the employed encodings.

\begin{proposition}
Algorithm~\ref{alg:aba-preferred-skept} decides skeptical acceptance under preferred semantics for ABA frameworks, i.e.,
for a given ABA framework $F=(\mathcal{L},\mathcal{R},\mathcal{A},\contraryempty)$ and 
sentence $s \in \mathcal{L}$, Algorithm~\ref{alg:aba-preferred-skept} 
returns YES if $s$ is skeptically accepted under preferred semantics in $F$, and NO otherwise.
\end{proposition}

\subsection{\abaplus Properties}
\label{sec:aba+prop}
We move on to \abaplus, in this section first giving an alternative characterization of $<$-admissible and $<$-complete semantics to better suit our algorithmic setting, and then showing complexity membership result for $<$-complete semantics.
In contrast to ABA (and AFs), credulous acceptance under $<$-admissible and $<$-complete semantics does not coincide~\cite[{{Example 3.6}}]{DBLP:phd/ethos/Cyras17}. 
Further, $<$-attacks differ from (non-preference-based) attacks, requiring more complex computation~\cite{LehtonenWJ:JAIR2021}.
We begin with stating conditions for an assumption set to be $<$-admissible or $<$-complete, on which we base our algorithms.
Define for an assumption set $A$ the set of assumptions $U$ not individually $<$-attacked by $A$ by $U = \{a \in \mathcal{A} \mid A$ does not $<$-attack $a\}$. 

\begin{proposition}
    \label{prop:abaplus-char}
    Given an \abaplus{} framework $F$, a conflict-free set of assumptions $A$ in $F$, and the set of assumptions $U$ that $A$ does not individually $<$-attack, it holds that 
    \begin{itemize}
        \item $A$ is $<$-admissible iff there is no set $B\subseteq U$ such that $A$ does not $<$-attack $B$ and $B$ $<$-attacks $A$, and 
        \item $A$ is $<$-complete iff $A$ is $<$-admissible and for all $a \in \mathcal{A} \setminus A$ it holds that $a$ is $<$-attacked by some $B \subseteq U$ such that $A$ does not $<$-attack $B$. 
    \end{itemize}
\begin{proof}
    For the first item, assume that $A$ is $<$-admissible. It follows that if a set $B$ of assumptions $<$-attacks $A$ we have $A$ $<$-attacks $B$ ($A$ defends itself due to admissibility). Thus, there is no $B$ s.t.\ $B$ $<$-attacks $A$ and $A$ does not $<$-attack $B$. 
    For the other direction, assume that there is no $B\subseteq U$ such that $A$ does not $<$-attack $B$ and $B$ $<$-attacks $A$. Suppose there is a set $C$ of assumptions that $<$-attacks $A$. If $C \cap (\mathcal{A}\setminus U) \not = \emptyset$ ($C$ contains an assumption outside $U$), then $A$ $<$-attacks $C$ (on a particular assumption, and, due to subset monotonicity of $<$-attacks, also $C$). Consider the case that $C \cap (\mathcal{A}\setminus U) = \emptyset$. Then $C \subseteq U$
    (since $U \subseteq \mathcal{A}$). If $A$ does not $<$-attack $C$, then we arrive at a contradiction (contradicts our assumption of the right hand side of the formal statement). Thus, $A$ $<$-attacks $C$. It follows that $A$ defends itself against all $C\subseteq\mathcal{A}$ that $<$-attack $A$. Since $A$ is assumed to be conflict-free, it follows that $A$ is $<$-admissible. 
    
    For the second item, assume that $A$ is $<$-complete. Then $A$ is $<$-admissible by definition. 
    Let $a \in \mathcal{A} \setminus A$. It follows from definition that $A$ does not defend $\{a\}$. This implies that there is a set $B$ that $<$-attacks $\{a\}$ and $A$ does not $<$-attack $B$. 
    Since $A$ $<$-attacks any set $C$ with $C\cap U \not = \emptyset$, we have $B \subseteq U$. 
    For the other direction, assume that $A$ is $<$-admissible, and for all $a \in \mathcal{A}\setminus A$ there is a set $B\subseteq U$ such that $B$ $<$-attacks $\{a\}$ and $A$ does not $<$-attack $B$. 
    Suppose $A$ is not $<$-complete: then there is a set $C$ such that $A$ defends $C$ and $C \not \subseteq A$. Let $c \in C \setminus A$, implying that $c\in \mathcal{A}\setminus A$. Then, by assumption, there is a $B\subseteq U$ such that $B$ $<$-attacks $\{c\}$ and $A$ does not $<$-attack $B$. 
This is a contradiction to $\{c\}$ being defended by $A$. Thus, $A$ is $<$-complete. 
\end{proof}
\end{proposition}

The first item implies that we can focus on assumption sets among $U$ for checking defense. 
From the second item, it follows that given a $<$-complete assumption set $A$, for every $a\in U$ (and even for every $a\in \mathcal{A})$ it holds that either $a\in A$, or $U <$-attacks $a$. This fact can be used to prune candidates for $<$-complete assumption sets.

Complementing earlier results, we show a complexity membership result for credulous acceptance under $<$-complete semantics. 
The proof uses Proposition~\ref{prop:abaplus-char} and an earlier result~\cite[{{Proposition 11}}]{LehtonenWJ:JAIR2021}: 
after guessing a set of assumptions, checking $<$-admissibility amounts to verifying whether each $<$-attacker is $<$-attacked (in coNP), and checking whether the set contains all defended sets amounts to checking for each individual assumption outside $A$ whether this assumption is not defended (each check in NP). 

\begin{theorem}
\label{thm:membership}
Credulous acceptance under $<$-complete semantics in \abaplus{} is in $\Sigma^P_2$. 
\begin{proof}
Non-deterministically construct a set of assumptions $A$. 
    Now check whether (i) the queried sentence is derivable from $A$, (ii) $A$ is $<$-admissible, and (iii) $A$ is $<$-complete. 
Checking (i) and conflict-freeness can be done in polynomial time. 
    Construct $U = \{ u \in \mathcal{A} \mid A \textnormal{ does not } <$-attack $u \}$, which is doable in polynomial time~\cite[{{Proposition 11, item 1}}]{LehtonenWJ:JAIR2021}.
    For checking further conditions of $<$-admissibility, check for each set of assumptions $B\subseteq U$ whether $B$ $<$-attacks $A$ without $A$ $<$-attacking $B$. Checking for two concrete sets of assumptions whether one $<$-attacks the other is doable in polynomial time~\cite[{{Proposition 11, items 1 and 2}}]{LehtonenWJ:JAIR2021}. Thus, one can check whether $A$ defends itself via a check in coNP. 
    For checking whether $A$ is also $<$-complete, check for each $a \in \mathcal{A} \setminus A$ whether there is some set $B\subseteq U$ such that $B$ $<$-attacks $a$ and $A$ does not $<$-attack $B$.
This is in NP. 
    These checks establish whether $A$ is $<$-complete, by Proposition~\ref{prop:abaplus-char}, and whether the queried sentence is derivable from $A$, satisfying the definition of credulous acceptance. 
    Overall, this gives a non-deterministic polynomial time algorithm that accesses an NP oracle, showing membership in $\Sigma^P_2$ for credulous acceptance under $<$-complete semantics.
\end{proof}
\end{theorem}

\subsection{$<$-Admissible Semantics}
\label{sec:adm-alg}

We proceed to algorithms for \abaplus, starting with Algorithm~\ref{alg:abap-adm-cred} for deciding credulous acceptance under $<$-admissible semantics in a given \abaplus{} framework $F$. This algorithm can be straightforwardly extended to cover enumeration of all $<$-admissible sets. We first give details of the algorithm, and subsequently explanations of the underlying ASP encodings. 

We represent a given \abaplus framework $F=(\mathcal{L},\mathcal{R},\mathcal{A},\contraryempty,\leq)$ in ASP as 
$    \mathtt{ABA^+}(F) = \mathtt{ ABA}(F)\cup \{ \asppreferred(x,y). \mid y \leq x \}\cup  \aspmodule{preferences}^+.$
Listing~\ref{asp:pref} shows $\aspmodule{preferences}^+$.
Algorithm~\ref{alg:abap-adm-cred} employs the ASP modules 
 $\pi_{\mathit{cand}} = \mathtt{ABA^+}(F) \cup \aspmodule{\cf}^+ \cup \{\mathit{constr}(\naf \aspsupported(s))\}$ and 
$\pi_{\mathit{check}} = \mathtt{ABA^+}(F) \cup \aspmodule{\mathit{defended}}^+ \cup \aspmodule{\mathit{suspect-defeat}}^+$. 
The former, $\pi_{\mathit{cand}}$, encodes the abstraction (candidate search space) by considering conflict-free sets of assumptions that contain the queried sentence $s$ in $\aspin(I)$, for an answer set $I$ of the encoding, and additionally computes all singleton assumptions not $<$-attacked by $\aspin(I)$, in the ASP atoms ${\bf undefeated}(I)$. 
In Line~\ref{alg:abap-adm-cred-5} of Algorithm~\ref{alg:abap-adm-cred} we check, based on Proposition~\ref{prop:abaplus-char}, whether $\aspin(I)$ corresponds to an $<$-admissible set in $F$: the ASP encoding is satisfiable iff there is a subset of ${\bf undefeated}(I)$ that is not $<$-attacked by ${\bf in}(I)$ but that $<$-attacks $\aspin(I)$ (via either normal or reverse $<$-attacks). 
If $\aspin(I)$ does correspond to an $<$-admissible set, this is a witness to $s$ being credulously accepted.
Otherwise, we exclude this assumption set via the constraint $\mathit{constr}(\aspout(I) \cup \aspin(I))$.

\begin{lstlisting}[caption={Module $\aspmodule{preferences}^+$},frame=lines,label=asp:pref]
preferred(X,Z) :- preferred(X,Y), preferred(Y,Z).
less_preferred(X,Y) :- preferred(Y,X), not preferred(X,Y).
no_less_preferred(X,Y) :- assumption(X), assumption(Y), not less_preferred(X,Y).
\end{lstlisting}

Algorithm~\ref{alg:abap-adm-cred} is extended to cover enumeration of $<$-admissible assumption sets by reporting all found $<$-admissible sets and not terminating until there are no more candidates; see~\ref{appendix:algs}.

\begin{algorithm}[!t]
\caption{Credulous acceptance under $<$-admissible semantics}
\label{alg:abap-adm-cred}
\begin{algorithmic}[1]
    \REQUIRE \abaplus{} framework $F=(\mathcal{L},\mathcal{R},\mathcal{A},\contraryempty, \leq)$
    \ENSURE return YES if $s$ is credulously accepted under $<$-admissible semantics in $F$, NO otherwise
    \STATE{$\pi_{\mathit{cand}} := \mathtt{ABA^+}(F) \cup \pi_{cf}\cup \aspmodule{\mathit{undefeated}}^+ \cup \{\mathit{constr}(\naf \aspsupported(s))\}$}
    \STATE{$\pi_{\mathit{check}} := \mathtt{ABA^+}(F) \cup \aspmodule{\mathit{defended}}^+ \cup \aspmodule{\mathit{suspect-defeat}}^+$}
    \STATE{$\algorithmicwhile\ \pi_{\mathit{cand}}$ is satisfiable $\algorithmicdo$}
    \STATE{\hspace{\algorithmicindent} Let $I$ be the found answer set}
    \STATE{\hspace{\algorithmicindent} $\algorithmicif\ \pi_{\mathit{check}} \cup {\bf undefeated}(I) \cup \aspin(I)$ is unsatisfiable $\algorithmicthen\ \algorithmicreturn$ YES} \label{alg:abap-adm-cred-5}
    \STATE{\hspace{\algorithmicindent} $\pi_{\mathit{cand}} := \pi_{\mathit{cand}} \cup \{\mathit{constr}(\aspout(I) \cup \aspin(I))\}$} 
    \RETURN{NO}
\end{algorithmic}
\end{algorithm}

\begin{lstlisting}[caption={Module $\aspmodule{\cf}^+$},frame=lines,label=asp:adm1]
in(X) :- assumption(X), not out(X).
out(X) :- assumption(X), not in(X).
supported(X) :- assumption(X), in(X).
supported(X) :- head(R,X), triggered_by_in(R).
triggered_by_in(R) :- head(R,_), supported(X) : body(R,X).
:- in(X), contrary(X,Y), supported(Y).
pref_supported(X,Y) :- no_less_preferred(X,Y), assumption(X), in(X).
pref_supported(X,Y) :- head(R,X), pref_triggered_by_in(R,Y).
pref_triggered_by_in(R,Y) :- head(R,_),assumption(Y),pref_supported(X,Y):body(R,X).
normally_defeated(Y) :- pref_supported(X,Y), contrary(Y,X).
derivable_from_undefeated(Z,Z) :- assumption(Z), not normally_defeated(Z).
derivable_from_undefeated(Y,Z) :- head(R,Y), triggered_by_undefeated(R,Z).
triggered_by_undefeated(R,Z) :- head(R,_),assumption(Z),&\newline&derivable_from_undefeated(Y,Z):body(R,Y).
in_attacked_by_normally_undefeated(X,Z):-in(X),contrary(X,Y),derivable_from_undefeated(Y,Z).
reversely_defeated(Z) :- less_preferred(Z,X), in_attacked_by_normally_undefeated(X,Z).
undefeated(X) :- assumption(X), not normally_defeated(X), not reversely_defeated(X).
\end{lstlisting}

\begin{lstlisting}[caption={Partial module $\aspmodule{defended}^+$},frame=lines,label=asp:adm2]
suspect(X) :- undefeated(X), not other(X).
other(X) :- undefeated(X), not suspect(X).
pref_supported_by_suspects(X,Y) :- in(Y), no_less_preferred(X,Y), assumption(X), suspect(X).
pref_supported_by_suspects(X,Y) :- in(Y), head(R,X), pref_triggered_by_suspects(R,Y).
pref_triggered_by_suspects(R,Y) :- in(Y), head(R,_), pref_supported_by_suspects(X,Y):body(R,X).
in_normally_defeated_by_suspects :- in(Y), pref_supported_by_suspects(X,Y), contrary(Y,X).
supported_by_in(X) :- assumption(X), in(X).
supported_by_in(X) :- head(R,X), triggered_by_in(R).
triggered_by_in(R) :- head(R,_), supported_by_in(X) : body(R,X).
reach_in(X,Y) :- triggered_by_in(R), head(R,Y), body(R,X).
reach_in(X,Y) :- reach_in(X,Z), reach_in(Z,Y).
reach_in(X,X) :- in(X).
in_reversely_defeated_by_suspects :- suspect(Y), contrary(Y,X), supported_by_in(X), in(Z),&\newline& reach_in(Z,X), less_preferred(Z,Y).
:- not in_normally_defeated_by_suspects, not in_reversely_defeated_by_suspects.
\end{lstlisting}

We present $\aspmodule{cf}^+$ in Listing~\ref{asp:adm1}, $\aspmodule{defended}^+$ in Listing~\ref{asp:adm2} and $\aspmodule{suspect-defeat}^+$ in Listing~\ref{asp:common}. 
The first six lines of Listing~\ref{asp:adm1} encode conflict-freeness (note that conflict-freeness is independent of preferences~\cite{CyrasT16-nmr}). 
In brief, $\aspin(I)$ encodes a guess of an assumption set and  $\aspsupported(I)$ which sentences are derivable from this set. 
For computing assumptions $x$ that are individually $<$-attacked by the assumptions $A$ encoded by $\aspin(I)$, we make use of a result proven by~\citeN[{{Lemma 8}}]{LehtonenWJ:JAIR2021}. 
Checking whether $A$ normally $<$-attacks an $x$ can directly be encoded by forward-derivations: if from the subset $A'\subseteq A$ which are not less preferred to $x$ one can derive the contrary of $x$, a normal $<$-attack from $A$ to $x$ exists. For the remaining assumptions, it holds that $A$ reversely $<$-attacks $x$ if from $\{x\}$ one can derive the contrary of an assumption $a \in A$ with $x< a$. 

From $\aspin(I)$ and ${\bf undefeated}(I)$ obtained as facts from an earlier ASP call, the encodings in Listings~\ref{asp:adm2} and~\ref{asp:common} determine whether $\aspin(I)$ defends itself against ${\bf undefeated}(I)$.
In Listing~\ref{asp:adm2}, we guess a subset of the undefeated assumptions (called suspects here) and check whether this set $<$-attacks $\aspin(I)$ but is not $<$-attacked by $\aspin(I)$ (making $\aspin(I)$ undefended and not $<$-admissible). Line~1 encodes the guess and normal $<$-attacks as before. Reverse $<$-attacks from a set larger than one are more involved; the idea is taken from~\citeN[{{proof of Proposition 11.2}}]{LehtonenWJ:JAIR2021}. We compute what is supported by
$\aspin(I)$.
The set $\aspin(I)$ is reversely $<$-attacked by the ${\bf suspect}(I)$ set if one can tree-derive from $\aspin(I)$ a contrary of an assumption $x$ in the ${\bf suspect}(I)$ set, with the required assumptions among $\aspin(I)$ having an assumption less preferred than $x$. To show that there is such a derivation tree from a subset $A$ of the assumptions corresponding to $\aspin(I)$ that derives a contrary $y$ of $x$, with one assumption $a \in A$ being less preferred than $x$, we check whether one can reach $y$ from $a$ via the derivation rules (implying existence of such a tree). 
In Listing~\ref{asp:common}, normal and reverse $<$-attacks from $\aspin(I)$ to \textbf{suspect}$(I)$ are determined analogously to $<$-attacks from \textbf{suspect}$(I)$ to $\aspin(I)$. 
The final constraints of Listings~\ref{asp:adm2} and~\ref{asp:common} ensure that \textbf{suspect}$(I)$ $<$-attacks $\aspin(I)$, but not vice versa.
If $\mathtt{ABA^+}(F) \cup \aspmodule{\mathit{defended}}^+ \cup \aspmodule{\mathit{suspect-defeat}}^+$ is unsatisfiable, the assumption set encoded in $\aspin(I)$ defends itself.

\begin{lstlisting}[caption={Module $\aspmodule{suspect-defeat}^+$},frame=lines,label=asp:common]
supported_by_in(X,Y) :- suspect(Y), no_less_preferred(X,Y), assumption(X), in(X).
supported_by_in(X,Y) :- suspect(Y), head(R,X), triggered_by_in(R,Y).
triggered_by_in(R,Y) :- suspect(Y), head(R,_), assumption(Y), supported_by_in(X,Y) : body(R,X).
suspect_normally_defeated_by_in :- supported_by_in(X,Y), contrary(Y,X).
supported_by_suspects(X) :- assumption(X), suspect(X).
supported_by_suspects(X) :- head(R,X), triggered_by_suspects(R).
triggered_by_suspects(R) :- head(R,_), supported_by_suspects(X) : body(R,X).
reach_suspect(X,Y) :- triggered_by_suspects(R), head(R,Y), body(R,X).
reach_suspect(X,Y) :- reach_suspect(X,Z), reach_suspect(Z,Y).
reach_suspect(X,X) :- suspect(X).
suspect_reversely_defeated_by_in :- in(Y), contrary(Y,X), supported_by_suspects(X), 
     suspect(Z), (Z,X), less_preferred(Z,Y).
:- suspect_normally_defeated_by_in.
:- suspect_reversely_defeated_by_in.
\end{lstlisting}

The following proposition states the correctness of Algorithm~\ref{alg:abap-adm-cred} based on 
Proposition~\ref{prop:abaplus-char} and the previous discussion on the details of the algorithm and the employed encodings.

\pagebreak
\begin{proposition}
    Algorithm~\ref{alg:abap-adm-cred} decides credulous acceptance under $<$-admissible semantics, i.e., for a given ABA framework $F=(\mathcal{L},\mathcal{R},\mathcal{A},\contraryempty)$ and $s \in \mathcal{L}$, Algorithm~\ref{alg:abap-adm-cred} returns YES if $s$ is credulously accepted in $F$, and NO otherwise. 
\end{proposition}

\subsection{$<$-Complete Semantics}
\label{sec:com-alg}

\begin{algorithm}[!t]
\caption{Credulous acceptance under $<$-complete semantics}
\label{alg:abap-com-cred}
\begin{algorithmic}[1]
    \REQUIRE \abaplus{} framework $F=(\mathcal{L},\mathcal{R},\mathcal{A},\contraryempty, \leq)$, $s\in \mathcal{L}$
    \ENSURE return YES if $s$ is credulously accepted under $<$-complete semantics in $F$, NO otherwise
    \STATE{$\pi_{\mathit{cand}} := \mathtt{ABA^+}(F) \cup \pi_{cf}\cup \aspmodule{\mathit{undefeated}}^+ \cup \aspmodule{prune}^+ \cup \{\mathit{constr}(\naf \aspsupported(s))\}$}
    \STATE{$\pi_{\mathit{check1}} := \mathtt{ABA^+}(F) \cup \aspmodule{defended}^+ \cup \aspmodule{suspect-defeat}^+$}
    \STATE{$\pi_{\mathit{check2}} := \mathtt{ABA^+}(F) \cup \aspmodule{com}^+ \cup \aspmodule{suspect-defeat}^+$}
    \STATE{$\algorithmicwhile\ \pi_{\mathit{cand}}$ is satisfiable $\algorithmicdo$}
    \STATE{\hspace{\algorithmicindent} Let $I$ be the found answer set; $\mathit{flag}:=true$}
    \STATE{\hspace{\algorithmicindent} $\algorithmicif\ \pi_{\mathit{check1}} \cup {\bf undefeated}(I) \cup \aspin(I)$ unsatisfiable $\algorithmicthen$}
    \STATE{\hspace{\algorithmicindent}\hspace{\algorithmicindent}$\algorithmicfor$ each $u\in \mathcal{A}$ such that ${\bf undefeated}(a)\in I\ \algorithmicdo$}
\STATE{\hspace{\algorithmicindent}\hspace{\algorithmicindent}\hspace{\algorithmicindent} $\algorithmicif\ \pi_{\mathit{check2}}
        \cup {\bf undefeated}(I)
        \cup \{{\bf target}(u)\}
        \cup {\bf in}(I)$ is\\\hspace{\algorithmicindent}\hspace{\algorithmicindent}\hspace{\algorithmicindent} unsatisfiable $\algorithmicthen\ \mathit{flag}:=false$; break}
        \STATE{\hspace{\algorithmicindent}\hspace{\algorithmicindent}$\algorithmicif\ \mathit{flag}=true\ \algorithmicthen\ \algorithmicreturn$ YES}
    \STATE{\hspace{\algorithmicindent} $\pi_{\mathit{cand}} := \pi_{\mathit{cand}} \cup \{\mathit{constr}(\aspout(I) \cup \aspin(I))\}$}
    \RETURN{NO}
\end{algorithmic}
\end{algorithm}

For deciding credulous acceptance under $<$-complete semantics, we present Algorithm~\ref{alg:abap-com-cred}. 
There are two key differences to Algorithm~\ref{alg:abap-adm-cred}: the abstraction $\pi_{\mathit{cand}}$ is stronger and verifying whether a candidate assumption set is $<$-complete is more involved. 
For the former, in addition to the constraints posed in Algorithm~\ref{alg:abap-adm-cred}, we add that if $\aspin(I)$ corresponds to a conflict-free set of assumptions $A$ and $U$ is the set of all assumptions which are not individually $<$-attacked by $A$, then for each $a \in U$ it must hold that either $a\in A$ or $a$ is $<$-attacked by $U$. By Proposition~\ref{prop:abaplus-char}, this only rules out assumption sets that are not $<$-complete. 
As we will see in the experiments, pruning the search space in this manner can significantly speed up computation. 
For verifying whether a conflict-free set of assumptions is $<$-complete, Algorithm~\ref{alg:abap-com-cred} checks in Lines~7--9 for each $a \in U$ whether $a$ is defended by $A$, in addition to verifying $<$-admissibility (Line~6). 

\begin{lstlisting}[caption={Module $\aspmodule{prune}^+$},frame=lines,label=asp:com1,float]
pref_supported_by_undefeated(X,Y) :- no_less_preferred(X,Y), assumption(X), undefeated(X).
pref_supported_by_undefeated(X,Y) :- head(R,X), pref_triggered_by_undefeated(R,Y).
pref_triggered_by_undefeated(R,Y) :- pref_supported_by_undefeated(X,Y) : body(R,X), 
     assumption(Y), head(R,_). 
undefeated_normally_defeated_by_undefeated(Y) :- undefeated(Y), &\newline&pref_supported_by_undefeated(X,Y), contrary(Y,X).
undefeated_reversely_defeated_by_undefeated(Z) :- less_preferred(Z,X), undefeated(X), 
     contrary(X,Y), derivable_from_undefeated(Y,Z).
:- out(Y), undefeated(Y), not undefeated_normally_defeated_by_undefeated(Y), not undefeated_reversely_defeated_by_undefeated(Y).
\end{lstlisting}

\begin{lstlisting}[caption={Module $\aspmodule{com}^+$},frame=lines,label=asp:com2,float]
suspect(X) :- assumption(X), not other(X).
other(X) :- assumption(X), not suspect(X).
pref_supported_by_suspects(X) :- target(Y),no_less_preferred(X,Y),assumption(X),suspect(X).
pref_supported_by_suspects(X) :- head(R,X), pref_triggered_by_suspects(R).
pref_triggered_by_suspects(R) :- head(R,_), pref_supported_by_suspects(X) : body(R,X).
target_normally_attacked :- target(Y), pref_supported_by_suspects(X), contrary(Y,X).
derivable_from_target(X) :- target(X).
derivable_from_target(X) :- head(R,X), triggered_by_target(R).
triggered_by_target(R) :- head(R,_), derivable_from_target(X) : body(R,X).
suspect_attacked_by_target(X) :- suspect(X), contrary(X,Y), derivable_from_target(Y).
target_reversely_attacked :- target(Y),less_preferred(Y,X),suspect_attacked_by_target(X).
:- not target_normally_attacked, not target_reversely_attacked.
\end{lstlisting}

Enumeration of $<$-complete assumption sets can be achieved by reporting all found answers and not terminating until there are no candidates, and finding a $<$-complete assumption set by omitting the query and
reporting $\aspin(I)$ on Line 9; see~\ref{appendix:algs}.
Enumeration also finds the $<$-grounded assumption set, which is defined as the intersection of all $<$-complete sets.

In module $\aspmodule{prune}^+$ (Listing~\ref{asp:com1}) we compute $<$-attacks on singleton assumptions. We consider singleton assumptions in $U$, and whether they are $<$-attacked by $U$. The final constraint rules out exactly the condition mentioned after Proposition~\ref{prop:abaplus-char}, namely that there is an $a\in \mathcal{A}$ such that $a\notin \aspin(I)$ and $a$ is not $<$-attacked by $U$. 

Via the encoding in Listing~\ref{asp:com2} together with Listing~\ref{asp:common} we check whether, given a set $\aspin(I)$ and an assumption ${\bf target}(I)$, $\aspin(I)$ defends ${\bf target}(I)$.
More specifically the module is unsatisfiable if there is no set of assumptions, called suspects here, that attack ${\bf target}(I)$ without $\aspin(I)$ attacking the suspect set.
In other words, the encoding is unsatisfiable if the ${\bf target}(I)$ is defended by $\aspin(I)$.
    We check if the target is normally or reversely $<$-attacked via \textbf{suspect}$(I)$. 
As before, in Listing~\ref{asp:common} we compute normal and reverse $<$-attacks from $\aspin(I)$ to the \textbf{suspect}$(I)$ set.

The following proposition states the correctness of the approach.

\begin{proposition}
    Algorithm~\ref{alg:abap-com-cred} decides credulous acceptance under $<$-complete semantics, i.e., for a given ABA framework $F=(\mathcal{L},\mathcal{R},\mathcal{A},\contraryempty)$ and $s \in \mathcal{L}$, Algorithm~\ref{alg:abap-com-cred} returns YES if $s$ is credulously accepted under $<$-complete semantics in $F$, and NO otherwise.
\end{proposition}

\section{Empirical Evaluation}

We implemented the ASP-based CEGAR algorithms using the incremental Python interface of Clingo v5.4.0~\cite{GebserKKOSW16,DBLP:journals/corr/abs-2008-06692}.
The implementation is available at \url{https://bitbucket.org/coreo-group/aspforaba}.
We empirically evaluate its performance, comparing it to 
current state-of-the-art approaches: the Asprin-based~\cite{BrewkaD0S15} approach to skeptical 
acceptance under preferred semantics~\cite{LehtonenWJ:JAIR2021} and the ABAplus system which supports computing assumption sets in ABA$^+$ under
$<$-admissible and $<$-complete semantics~\cite{BaoCT17}. A direct comparison with ABAplus is only applicable for
frameworks which satisfy the so-called WCP property due to restrictions in ABAplus.

For comparison with Asprin 
we use similar benchmarks as~\citeN{LehtonenWJ:JAIR2021}, with
$|\mathcal{L}|= 250, 500, 1000, 1500, \ldots, 8000$.
For each $|\mathcal{L}|$, we generated 20 frameworks with 15\% and 30\% of the sentences assumptions each for a total of 680 frameworks.
The number of rules per head and body lengths, respectively, were randomly chosen from $[1,20]$. 
For comparison with ABAplus, we use the 120 frameworks that satisfy the WCP property 
first used by~\citeN{LehtonenWJ:JAIR2021} containing up to 30 sentences.
The experiments were run  single-threaded on 2.6-GHz Intel Xeon E5-2670 processors
using per-instance  600-s time and 16-GB memory limit.

\begin{figure}[!b]
\centering
        \includegraphics[width=0.48\linewidth]{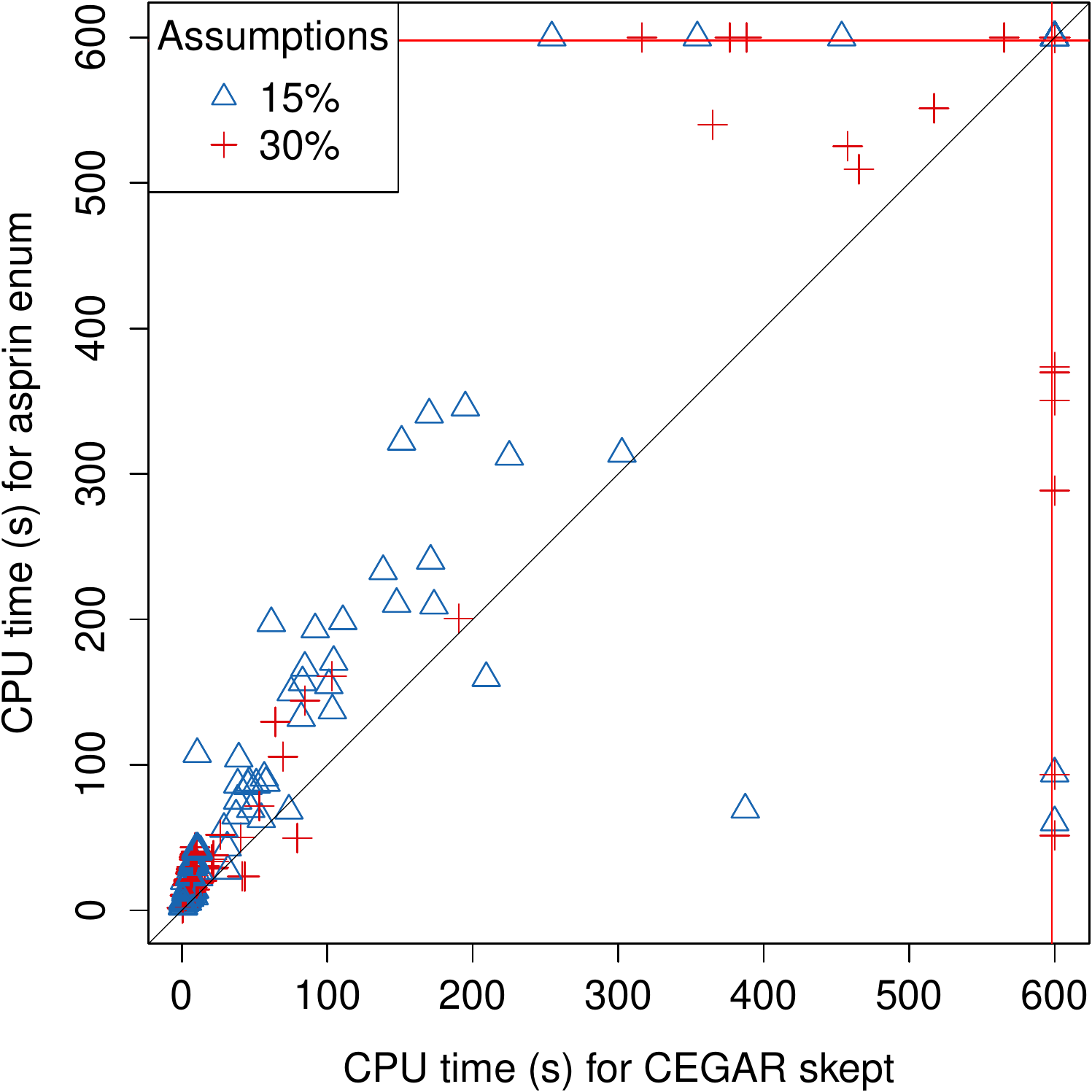}
        \hspace{5pt}
    \includegraphics[width=0.48\linewidth]{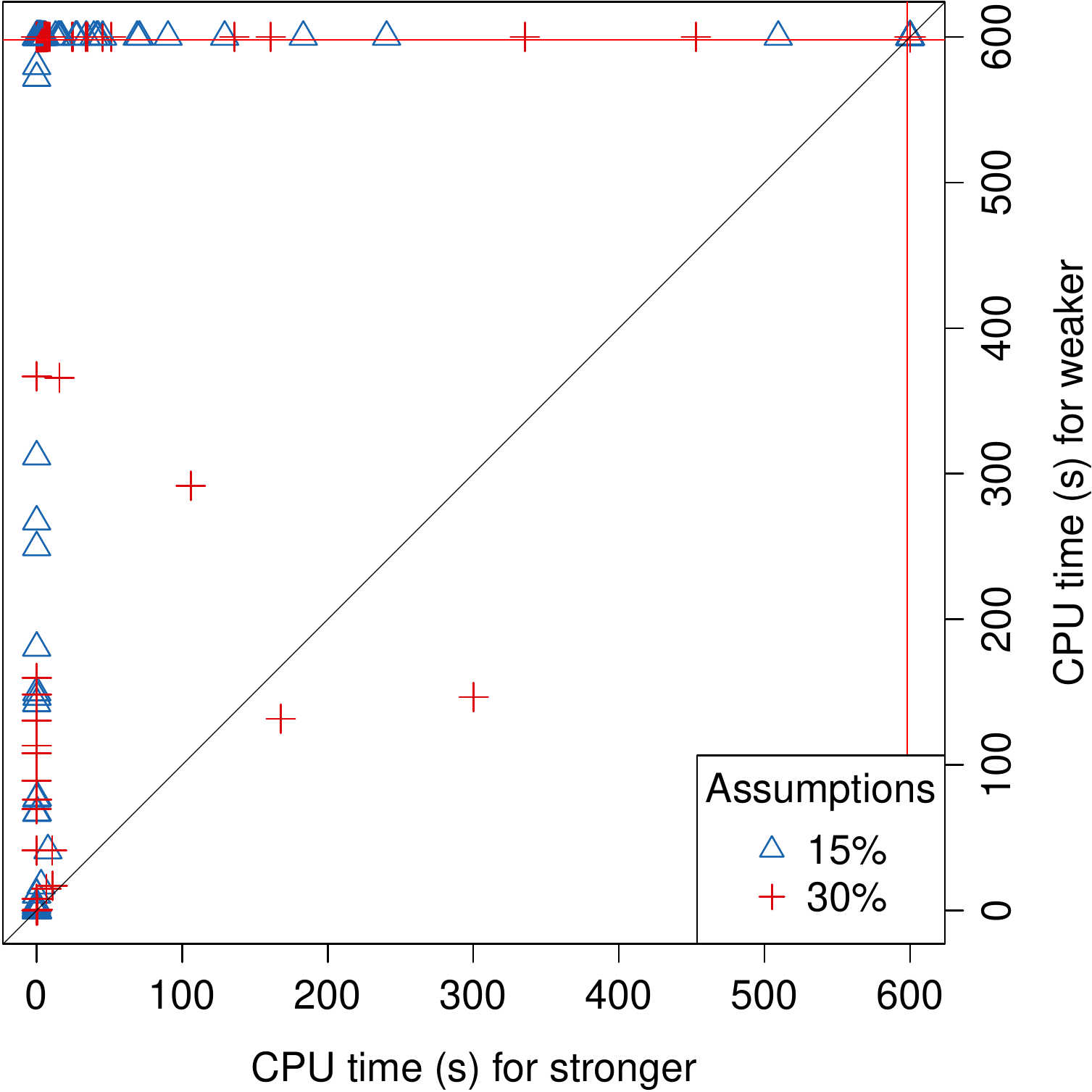}
    \caption{Left: Runtime comparison of Asprin (enumeration) and incremental ASP on skeptical preferred.
    Right: Runtime comparison of the incremental ASP approach using the weaker and stronger abstraction for finding an assumption set under $<$-complete semantics.}
\label{fig:primary}
\end{figure}

Figure~\ref{fig:primary} (left) shows a per-instance runtime comparison of the Asprin-based approach 
and our incremental ASP-based CEGAR algorithm for skeptical ABA reasoning under preferred semantics. 
The CEGAR approach clearly outperforms Asprin.
A comparison of ABAplus and our CEGAR approach is shown in Table~\ref{table:wcp}. 
The CEGAR approach dominates ABAplus in performance on the task of assumption set enumeration (as supported by ABAplus)
under both $<$-admissible and $<$-complete semantics. We conclude that the CEGAR algorithms based on incremental ASP outperform the current state of the art
on all of the three reasoning tasks.

\begin{figure}
\centering
        \includegraphics[width=0.48\linewidth]{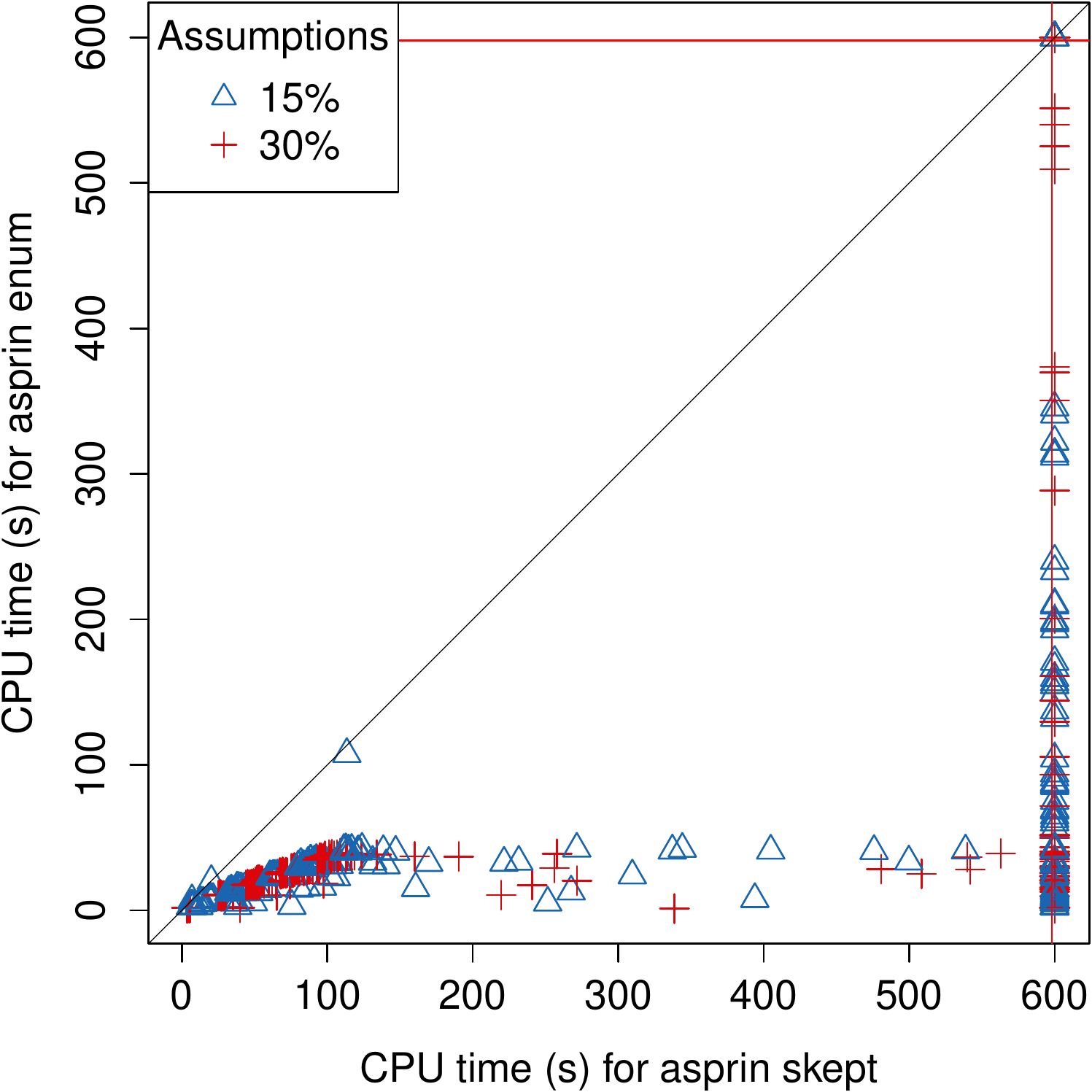}
        \hspace{5pt}
        \includegraphics[width=.48\linewidth]{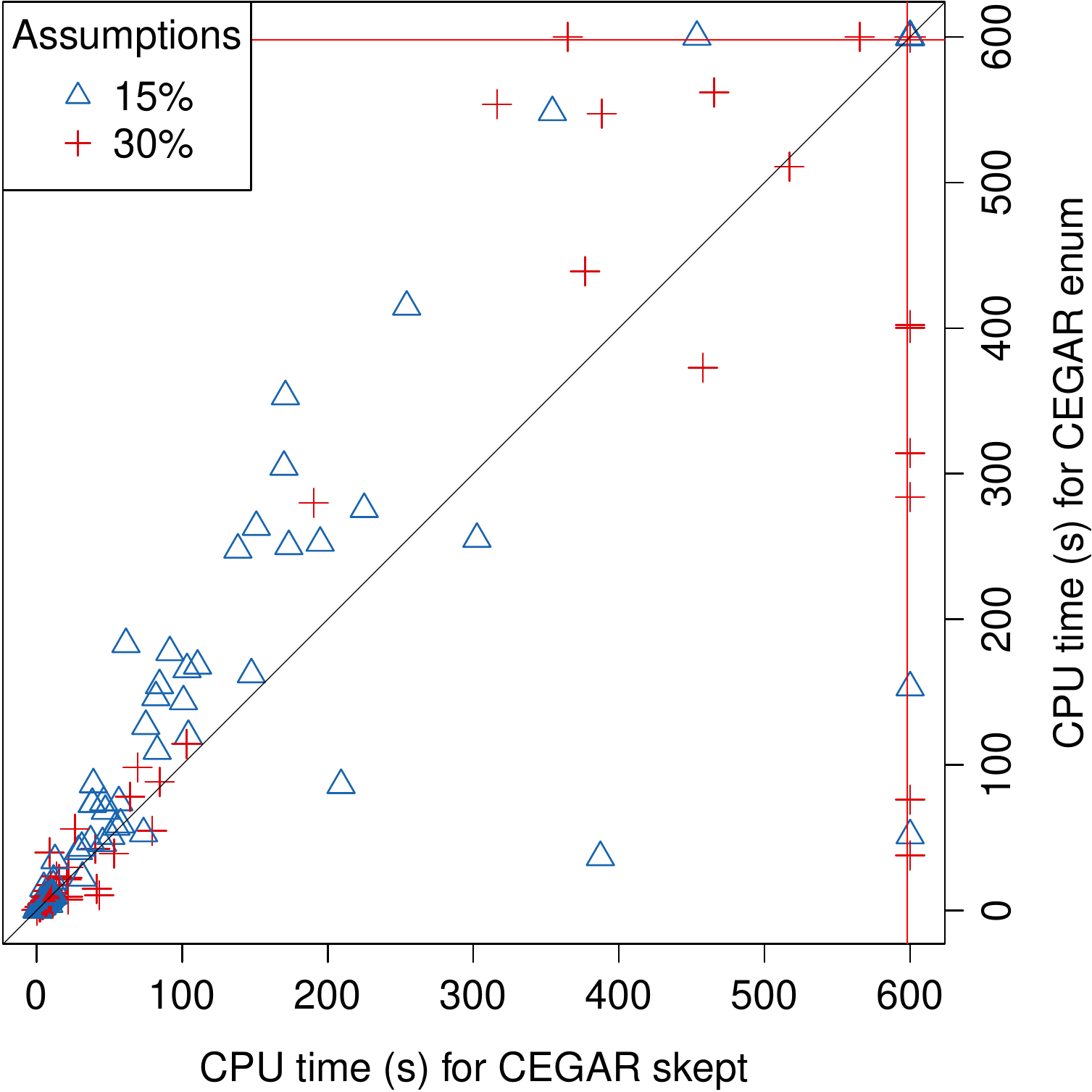}
        \caption{Runtime comparisons under preferred semantics. Left: Asprin enumeration vs skeptical reasoning.
            Right: CEGAR enumeration vs skeptical reasoning.}
\label{fig:prf-extra}
\end{figure}

Runtimes for Asprin are in its enumeration mode rather than query mode as Asprin is consistently faster on this task using enumeration, as shown in Figure~\ref{fig:prf-extra} (left). 
Conversely, Figure~\ref{fig:prf-extra} (right) shows that using our CEGAR approach, most instances are solved faster via direct skeptical reasoning compared to assumption set enumeration; 
there is only a handful of instances on which enumeration is faster.

\begin{figure}[!b]
\centering
    \includegraphics[width=0.48\linewidth]{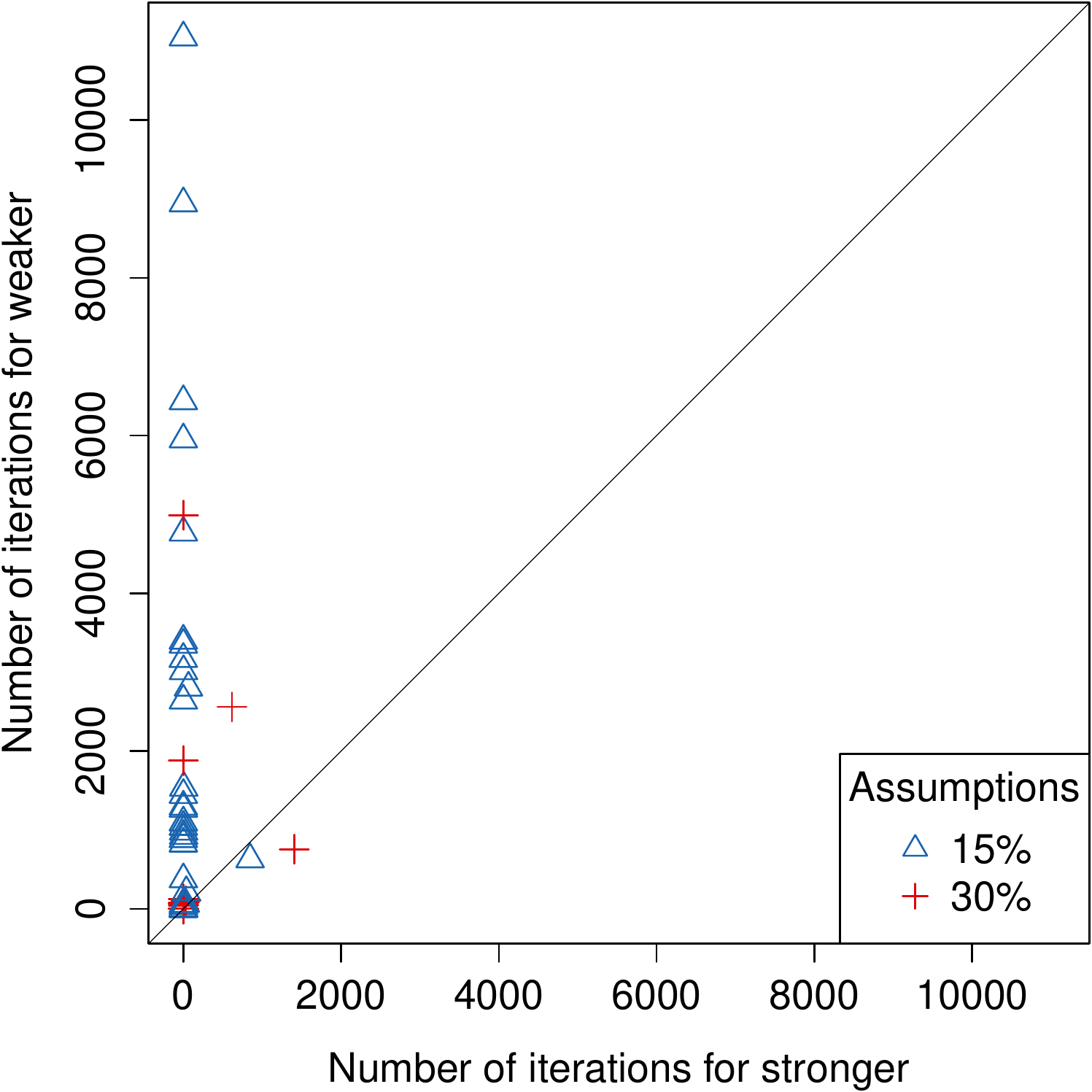}
        \hspace{5pt}
    \includegraphics[width=0.48\linewidth]{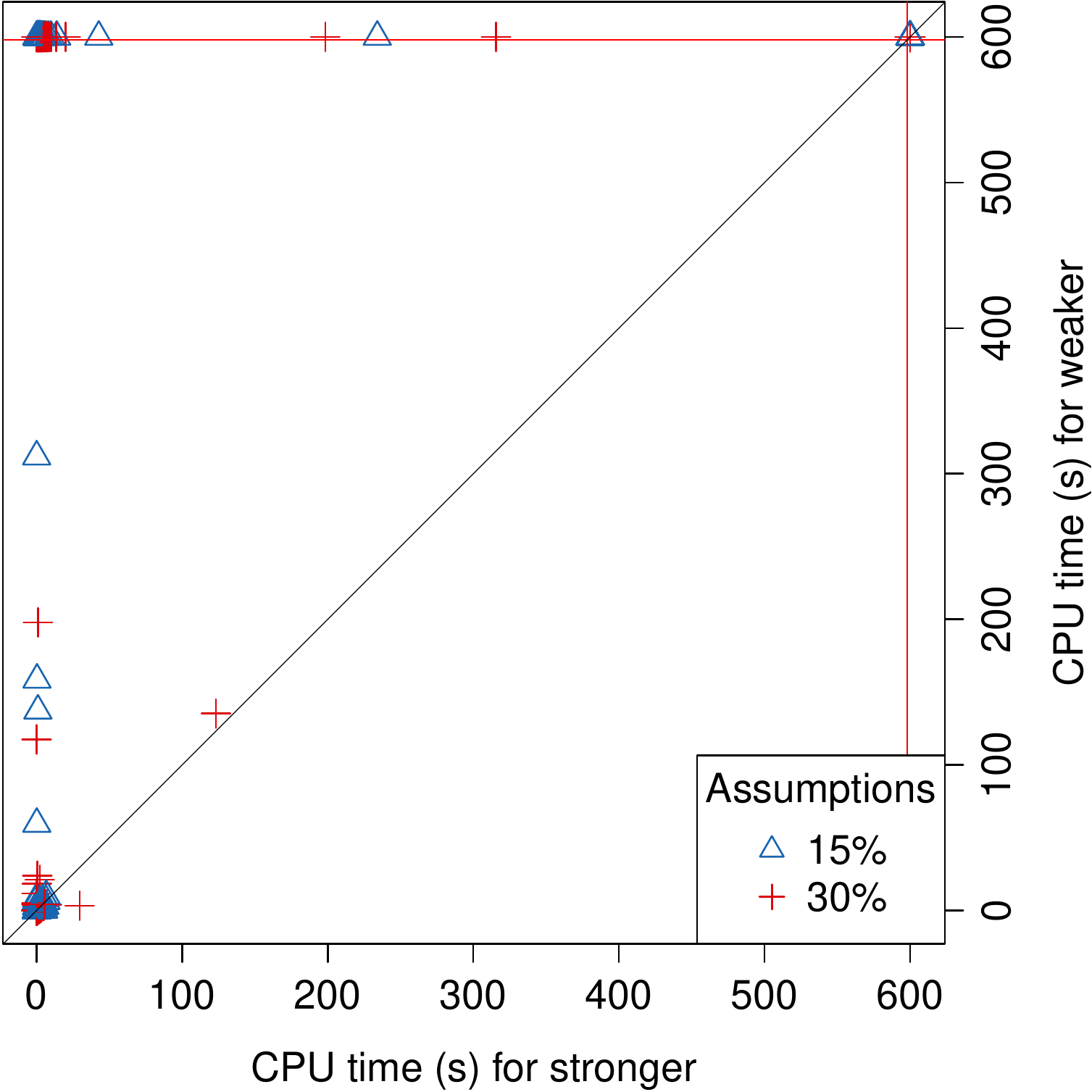}
    \caption{Comparisons for $<$-complete semantics using the CEGAR approach. Left: Comparison of iterations needed using the weaker and stronger abstraction for finding a $<$-complete assumption set.
    Right: Runtime comparison between the weaker and stronger abstraction for answering credulous acceptance.}
\label{fig:com-extra}
\end{figure}

For more insights into our CEGAR approach to ABA$^+$,
we  generated larger instances with
50-500 sentences.
For each $|\mathcal{L}|$, we generated 30 instances with 15\% and 30\% assumptions each
taking as preferences
random permutations of the assumptions, with assumption $a_i$ set to be preferred to $a_j$ for $i<j$ in the permutation
with probabilities 5\%, 15\% or 40\% (10 instances for each probability).
On these instances 
the stronger abstraction for $<$-complete semantics yields significant runtime improvements over the weaker abstraction, enabling scaling
up to 500 sentences (Figure~\ref{fig:primary} right).
The runtime improvements are at least in part due to the fact that the stronger abstraction results in considerably fewer iterations (essentially number of candidates found).
Using the weaker abstraction the algorithm takes on average 1148 iterations, compared to 12 when using the stronger abstraction.
Figure~\ref{fig:com-extra} (left) shows the iterations taken to solve each instance.
Further, the runtimes of the CEGAR approach using the stronger abstraction under $<$-complete semantics are similar between the task of finding an assumption set without a query and credulous reasoning on both unsatisfiable and satisfiable instances; Figure~\ref{fig:com-extra} (right) shows the overall runtime results for credulous reasoning under $<$-complete semantics.
We also observe that with a larger number of sentences being assumptions (30\% vs 15\%) 
instances tend to become harder to solve for all the considered problems.

\begin{table}[!t]
\caption{Runtime comparison.
Cumulative runtimes are over solved instances. \label{table:wcp}}
\centering
\begin{tabular}{@{}llrrrr@{ }}
\multicolumn{3}{c}{} & \multicolumn{3}{c}{\textbf{Running times} (s)}\\
\textbf{Problem} & \textbf{Approach} & \textbf{\#timeouts} & \textbf{mean}  & \textbf{median}  & \textbf{cumulative}\\
\hline
\abaplus{} $<$-$\adm$ & ASP & \textbf{0} & \textbf{0.114} & \textbf{0.040} & \textbf{14}\\
\emph{enumeration} & ABAplus & 9  &  15.442 & 0.560 & 1714\\
\hline
\abaplus{} $<$-$\comp$ & ASP & \textbf{0} & \textbf{0.096} & \textbf{0.040} & \textbf{12}\\
\emph{enumeration} & ABAplus & 9  &  14.240 & 0.550 & 1581\\
\hline
\end{tabular}
\end{table}

\section{Conclusions}
We developed an approach to beyond-NP reasoning in assumption-based argumentation frameworks
based on recent advances in incremental answer set solving. In particular, 
we detailed ASP-based counterexample-guided abstraction refinement procedures
for skeptical acceptance under preferred semantics in ABA and credulous reasoning under $<$-admissible and $<$-complete semantics in ABA$^+$, and assumption set enumeration for all of these. 
Our implementation of the approach
empirically outperforms previous algorithmic solutions to these reasoning tasks. 
We developed a stricter abstraction for $<$-complete semantics, speeding up solving in practice, and
obtained complexity upper bounds for credulous reasoning in ABA$^+$ under $<$-complete semantics.
A promising direction for further work is to extend the CEGAR approach considered in this work to other beyond-NP reasoning problems in ABA, such as reasoning over general (i.e. possibly non-flat) ABA frameworks.


\bibliographystyle{acmtrans}
\bibliography{references}

\begin{thebibliography}{}

\bibitem[\protect\citeauthoryear{Bao, {\v C}yras, and Toni}{Bao
  et~al\mbox{.}}{2017}]{BaoCT17}
{\sc Bao, Z.}, {\sc {\v C}yras, K.}, {\sc and} {\sc Toni, F.} 2017.
\newblock {ABAplus}: Attack reversal in abstract and structured argumentation
  with preferences.
\newblock In {\em Proc.~PRIMA}. LNCS, vol. 10621. Springer, 420--437.

\bibitem[\protect\citeauthoryear{Baroni, Gabbay, Giacomin, and van~der
  Torre}{Baroni et~al\mbox{.}}{2018}]{arguHandbook}
{\sc Baroni, P.}, {\sc Gabbay, D.}, {\sc Giacomin, M.}, {\sc and} {\sc van~der
  Torre, L.}, Eds. 2018.
\newblock {\em Handbook of Formal Argumentation}.
\newblock College Publications.

\bibitem[\protect\citeauthoryear{Besnard and Hunter}{Besnard and
  Hunter}{2018}]{BesnardH18}
{\sc Besnard, P.} {\sc and} {\sc Hunter, A.} 2018.
\newblock A review of argumentation based on deductive arguments.
\newblock In {\em Handbook of Formal Argumentation}. College Publications,
  Chapter~9, 437--484.

\bibitem[\protect\citeauthoryear{Bondarenko, Dung, Kowalski, and
  Toni}{Bondarenko et~al\mbox{.}}{1997}]{BondarenkoDKT97}
{\sc Bondarenko, A.}, {\sc Dung, P.~M.}, {\sc Kowalski, R.~A.}, {\sc and} {\sc
  Toni, F.} 1997.
\newblock An abstract, argumentation-theoretic approach to default reasoning.
\newblock {\em Artificial Intelligence\/}~{\em 93}, 63--101.

\bibitem[\protect\citeauthoryear{Brewka, Delgrande, Romero, and Schaub}{Brewka
  et~al\mbox{.}}{2015}]{BrewkaD0S15}
{\sc Brewka, G.}, {\sc Delgrande, J.~P.}, {\sc Romero, J.}, {\sc and} {\sc
  Schaub, T.} 2015.
\newblock asprin: Customizing answer set preferences without a headache.
\newblock In {\em Proc.~AAAI}. {AAAI} Press, 1467--1474.

\bibitem[\protect\citeauthoryear{Caminada and Schulz}{Caminada and
  Schulz}{2017}]{CaminadaS17}
{\sc Caminada, M.} {\sc and} {\sc Schulz, C.} 2017.
\newblock On the equivalence between assumption-based argumentation and logic
  programming.
\newblock {\em Journal of Artificial Intelligence Research\/}~{\em 60},
  779--825.

\bibitem[\protect\citeauthoryear{Cerutti, Gaggl, Thimm, and Wallner}{Cerutti
  et~al\mbox{.}}{2018}]{CeruttiGTW2018}
{\sc Cerutti, F.}, {\sc Gaggl, S.~A.}, {\sc Thimm, M.}, {\sc and} {\sc Wallner,
  J.~P.} 2018.
\newblock Foundations of implementations for formal argumentation.
\newblock In {\em Handbook of Formal Argumentation}. College Publications,
  Chapter~15, 688--767.

\bibitem[\protect\citeauthoryear{Clarke, Grumberg, Jha, Lu, and Veith}{Clarke
  et~al\mbox{.}}{2003}]{DBLP:journals/jacm/ClarkeGJLV03}
{\sc Clarke, E.~M.}, {\sc Grumberg, O.}, {\sc Jha, S.}, {\sc Lu, Y.}, {\sc and}
  {\sc Veith, H.} 2003.
\newblock Counterexample-guided abstraction refinement for symbolic model
  checking.
\newblock {\em Journal of the {ACM}\/}~{\em 50,\/}~5, 752--794.

\bibitem[\protect\citeauthoryear{Clarke, Gupta, and Strichman}{Clarke
  et~al\mbox{.}}{2004}]{DBLP:journals/tcad/ClarkeGS04}
{\sc Clarke, E.~M.}, {\sc Gupta, A.}, {\sc and} {\sc Strichman, O.} 2004.
\newblock {SAT}-based counterexample-guided abstraction refinement.
\newblock {\em {IEEE} Transactions on Computer Aided Design of Integrated
  Circuits and Systems\/}~{\em 23,\/}~7, 1113--1123.

\bibitem[\protect\citeauthoryear{Craven and Toni}{Craven and
  Toni}{2016}]{CravenT16}
{\sc Craven, R.} {\sc and} {\sc Toni, F.} 2016.
\newblock Argument graphs and assumption-based argumentation.
\newblock {\em Artificial Intelligence\/}~{\em 233}, 1--59.

\bibitem[\protect\citeauthoryear{Craven, Toni, and Williams}{Craven
  et~al\mbox{.}}{2013}]{CravenTW13}
{\sc Craven, R.}, {\sc Toni, F.}, {\sc and} {\sc Williams, M.} 2013.
\newblock Graph-based dispute derivations in assumption-based argumentation.
\newblock In {\em TAFA 2013 Revised Selected Papers}. LNCS, vol. 8306.
  Springer, 46--62.

\bibitem[\protect\citeauthoryear{{\v C}yras}{{\v
  C}yras}{2017}]{DBLP:phd/ethos/Cyras17}
{\sc {\v C}yras, K.} 2017.
\newblock {ABA+}: assumption-based argumentation with preferences.
\newblock Ph.D. thesis, Imperial College London, {UK}.

\bibitem[\protect\citeauthoryear{{\v C}yras, Fan, Schulz, and Toni}{{\v C}yras
  et~al\mbox{.}}{2018}]{CyrasFST2018}
{\sc {\v C}yras, K.}, {\sc Fan, X.}, {\sc Schulz, C.}, {\sc and} {\sc Toni, F.}
  2018.
\newblock Assumption-based argumentation: Disputes, explanations, preferences.
\newblock In {\em Handbook of Formal Argumentation}. College Publications,
  Chapter~7, 365--408.

\bibitem[\protect\citeauthoryear{{\v C}yras and Oliveira}{{\v C}yras and
  Oliveira}{2019}]{CyrasO19}
{\sc {\v C}yras, K.} {\sc and} {\sc Oliveira, T.} 2019.
\newblock Resolving conflicts in clinical guidelines using argumentation.
\newblock In {\em Proc.~{AAMAS}}. IFAAMAS, 1731--1739.

\bibitem[\protect\citeauthoryear{{\v C}yras and Toni}{{\v C}yras and
  Toni}{2016a}]{CyrasT16}
{\sc {\v C}yras, K.} {\sc and} {\sc Toni, F.} 2016a.
\newblock {ABA+}: Assumption-based argumentation with preferences.
\newblock In {\em Proc.~KR}. {AAAI} Press, 553--556.

\bibitem[\protect\citeauthoryear{{\v C}yras and Toni}{{\v C}yras and
  Toni}{2016b}]{CyrasT16-nmr}
{\sc {\v C}yras, K.} {\sc and} {\sc Toni, F.} 2016b.
\newblock Properties of {ABA+} for non-monotonic reasoning.
\newblock In {\em Proc.~NMR}. 25--34.

\bibitem[\protect\citeauthoryear{Dimopoulos, Nebel, and Toni}{Dimopoulos
  et~al\mbox{.}}{2002}]{DBLP:journals/ai/DimopoulosNT02}
{\sc Dimopoulos, Y.}, {\sc Nebel, B.}, {\sc and} {\sc Toni, F.} 2002.
\newblock On the computational complexity of assumption-based argumentation for
  default reasoning.
\newblock {\em Artificial Intelligence\/}~{\em 141,\/}~1/2, 57--78.

\bibitem[\protect\citeauthoryear{Dung, Kowalski, and Toni}{Dung
  et~al\mbox{.}}{2006}]{DungKT06}
{\sc Dung, P.~M.}, {\sc Kowalski, R.~A.}, {\sc and} {\sc Toni, F.} 2006.
\newblock Dialectic proof procedures for assumption-based, admissible
  argumentation.
\newblock {\em Artificial Intelligence\/}~{\em 170,\/}~2, 114--159.

\bibitem[\protect\citeauthoryear{Dung, Toni, and Mancarella}{Dung
  et~al\mbox{.}}{2010}]{DungTM10}
{\sc Dung, P.~M.}, {\sc Toni, F.}, {\sc and} {\sc Mancarella, P.} 2010.
\newblock Some design guidelines for practical argumentation systems.
\newblock In {\em Proc.~{COMMA}}. FAIA, vol. 216. {IOS} Press, 183--194.

\bibitem[\protect\citeauthoryear{Fan and Toni}{Fan and
  Toni}{2016}]{Fan:2016:IGA:2936924.2936964}
{\sc Fan, X.} {\sc and} {\sc Toni, F.} 2016.
\newblock On the interplay between games, argumentation and dialogues.
\newblock In {\em Proc.~AAMAS}. {ACM}, 260--268.

\bibitem[\protect\citeauthoryear{Fan, Toni, Mocanu, and Williams}{Fan
  et~al\mbox{.}}{2014}]{DBLP:conf/atal/FanTMW14}
{\sc Fan, X.}, {\sc Toni, F.}, {\sc Mocanu, A.}, {\sc and} {\sc Williams, M.}
  2014.
\newblock Dialogical two-agent decision making with assumption-based
  argumentation.
\newblock In {\em Proc.~AAMAS}. {IFAAMAS/ACM}, 533--540.

\bibitem[\protect\citeauthoryear{Gaertner and Toni}{Gaertner and
  Toni}{2007}]{GartnerT2007b}
{\sc Gaertner, D.} {\sc and} {\sc Toni, F.} 2007.
\newblock {CaSAPI}: A system for credulous and sceptical argumentation.
\newblock In {\em Proc.~NMR}. 80--95.

\bibitem[\protect\citeauthoryear{Garc{\'{\i}}a and Simari}{Garc{\'{\i}}a and
  Simari}{2018}]{GarciaS18}
{\sc Garc{\'{\i}}a, A.~J.} {\sc and} {\sc Simari, G.~R.} 2018.
\newblock Argumentation based on logic programming.
\newblock In {\em Handbook of Formal Argumentation}. College Publications,
  Chapter~8, 409--435.

\bibitem[\protect\citeauthoryear{Gebser, Kaminski, Kaufmann, Ostrowski, Schaub,
  and Wanko}{Gebser et~al\mbox{.}}{2016}]{GebserKKOSW16}
{\sc Gebser, M.}, {\sc Kaminski, R.}, {\sc Kaufmann, B.}, {\sc Ostrowski, M.},
  {\sc Schaub, T.}, {\sc and} {\sc Wanko, P.} 2016.
\newblock Theory solving made easy with {C}lingo 5.
\newblock In {\em Technical Communications of ICLP}. {OASICS}. Schloss Dagstuhl
  - Leibniz-Zentrum fuer Informatik, 2:1--2:15.

\bibitem[\protect\citeauthoryear{Gebser, Kaufmann, Kaminski, Ostrowski, Schaub,
  and Schneider}{Gebser
  et~al\mbox{.}}{2011}]{DBLP:journals/aicom/GebserKKOSS11}
{\sc Gebser, M.}, {\sc Kaufmann, B.}, {\sc Kaminski, R.}, {\sc Ostrowski, M.},
  {\sc Schaub, T.}, {\sc and} {\sc Schneider, M.~T.} 2011.
\newblock Potassco: The {P}otsdam answer set solving collection.
\newblock {\em {AI} Communications\/}~{\em 24,\/}~2, 107--124.

\bibitem[\protect\citeauthoryear{Gelfond and Lifschitz}{Gelfond and
  Lifschitz}{1988}]{DBLP:conf/iclp/GelfondL88}
{\sc Gelfond, M.} {\sc and} {\sc Lifschitz, V.} 1988.
\newblock The stable model semantics for logic programming.
\newblock In {\em Proc.~ ICLP/SLP}. {MIT} Press, 1070--1080.

\bibitem[\protect\citeauthoryear{Kaminski, Romero, Schaub, and Wanko}{Kaminski
  et~al\mbox{.}}{2020}]{DBLP:journals/corr/abs-2008-06692}
{\sc Kaminski, R.}, {\sc Romero, J.}, {\sc Schaub, T.}, {\sc and} {\sc Wanko,
  P.} 2020.
\newblock How to build your own {ASP}-based system?!
\newblock {\em CoRR\/}~{\em abs/2008.06692}.

\bibitem[\protect\citeauthoryear{Lehtonen, Wallner, and
  J{\"{a}}rvisalo}{Lehtonen et~al\mbox{.}}{2017}]{LehtonenWJ17}
{\sc Lehtonen, T.}, {\sc Wallner, J.~P.}, {\sc and} {\sc J{\"{a}}rvisalo, M.}
  2017.
\newblock From structured to abstract argumentation: Assumption-based
  acceptance via {AF} reasoning.
\newblock In {\em Proc.~ECSQARU}. LNCS, vol. 10369. Springer, 57--68.

\bibitem[\protect\citeauthoryear{Lehtonen, Wallner, and
  J{\"{a}}rvisalo}{Lehtonen et~al\mbox{.}}{2021}]{LehtonenWJ:JAIR2021}
{\sc Lehtonen, T.}, {\sc Wallner, J.~P.}, {\sc and} {\sc J{\"{a}}rvisalo, M.}
  2021.
\newblock Declarative algorithms and complexity results for assumption-based
  argumentation.
\newblock {\em Journal of Artificial Intelligence Research\/}~{\em 71},
  265--318.

\bibitem[\protect\citeauthoryear{Modgil and Prakken}{Modgil and
  Prakken}{2018}]{ModgilP18}
{\sc Modgil, S.} {\sc and} {\sc Prakken, H.} 2018.
\newblock Abstract rule-based argumentation.
\newblock In {\em Handbook of Formal Argumentation}. College Publications,
  Chapter~6, 287--364.

\bibitem[\protect\citeauthoryear{Niemel{\"{a}}}{Niemel{\"{a}}}{1999}]{DBLP:journals/amai/Niemela99}
{\sc Niemel{\"{a}}, I.} 1999.
\newblock Logic programs with stable model semantics as a constraint
  programming paradigm.
\newblock {\em Annals of Mathematics and Artificial Intelligence\/}~{\em
  25,\/}~3-4, 241--273.

\bibitem[\protect\citeauthoryear{Toni}{Toni}{2013}]{Toni13}
{\sc Toni, F.} 2013.
\newblock A generalised framework for dispute derivations in assumption-based
  argumentation.
\newblock {\em Artificial Intelligence\/}~{\em 195}, 1--43.

\bibitem[\protect\citeauthoryear{Toni}{Toni}{2014}]{Toni14}
{\sc Toni, F.} 2014.
\newblock A tutorial on assumption-based argumentation.
\newblock {\em Argument {\&} Computation\/}~{\em 5,\/}~1, 89--117.

\end{thebibliography}

\appendix

\section{Algorithms}
\label{appendix:algs}

We provide details on the variants of the algorithms presented in the main paper.

Algorithm~\ref{alg:aba-preferred-enum} enumerates preferred assumption sets, as a variant Algorithm~\ref{alg:aba-preferred-skept}.

\begin{algorithm}[h]
\caption{Assumption set enumeration under preferred semantics}
\label{alg:aba-preferred-enum}
\begin{algorithmic}[1]
\REQUIRE ABA framework $F=(\mathcal{L},\mathcal{R},\mathcal{A},\contraryempty)$
    \ENSURE return all preferred assumption sets of $F$
    \STATE{$\pi := \mathtt{ABA}(F) \cup \aspmodule{com}$}
    \STATE{$\algorithmicwhile\ \pi$ is satisfiable $\algorithmicdo$}
    \STATE{\hspace{\algorithmicindent} Let $I$ be the found answer set}
    \STATE{\hspace{\algorithmicindent} $\pi := \pi \cup \{\mathit{constr}(\aspout(I))\}$} 
    \STATE{\hspace{\algorithmicindent} $\algorithmicwhile\ \pi \cup \aspin(I)$ is satisfiable$\ \algorithmicdo$} 
    \STATE{\hspace{\algorithmicindent}\hspace{\algorithmicindent} Let $I$ be the found answer set}
    \STATE{\hspace{\algorithmicindent}\hspace{\algorithmicindent} $\pi :=\pi \cup \{\mathit{constr}(\aspout(I))\}$} 
    \STATE{\hspace{\algorithmicindent} $E := E\cup \{I\}$} 
    \RETURN $E$
\end{algorithmic}
\end{algorithm}

Algorithm~\ref{alg:abap-adm-enum} details an algorithm for $<$-admissible assumption set enumeration, as a variant of Algorithm~\ref{alg:abap-adm-cred} for deciding credulous acceptance under the same semantics. 

\begin{algorithm}
\caption{Assumption set enumeration under $<$-admissible semantics}
\label{alg:abap-adm-enum}
\begin{algorithmic}[1]
    \REQUIRE \abaplus{} framework $F=(\mathcal{L},\mathcal{R},\mathcal{A},\contraryempty, \leq)$
    \ENSURE return all $<$-admissible assumption sets of $F$
    \STATE{$\pi_{\mathit{cand}} := \mathtt{ABA^+}(F) \cup \pi_{cf}\cup \aspmodule{\mathit{undefeated}}^+$}
    \STATE{$\pi_{\mathit{check}} := \mathtt{ABA^+}(F) \cup \aspmodule{defended}^+ \cup \aspmodule{suspect-defeat}^+$}
    \STATE{$\algorithmicwhile\ \pi_{\mathit{cand}}$ is satisfiable $\algorithmicdo$}
    \STATE{\hspace{\algorithmicindent} Let $I$ be the found answer set}
    \STATE{\hspace{\algorithmicindent} $\algorithmicif\ \pi_{\mathit{check}} \cup {\bf undefeated}(I) \cup \aspin(I)$ is unsatisfiable $\algorithmicthen\ E := E\cup\{I\}$} 
    \STATE{\hspace{\algorithmicindent} $\pi_{\mathit{cand}} := \pi_{\mathit{cand}} \cup \{\mathit{constr}(\aspout(I) \cup \aspin(I))\}$} 
    \RETURN{$E$}
\end{algorithmic}
\end{algorithm}

Algorithm~\ref{alg:abap-com-find} is a variant of Algorithm~\ref{alg:abap-com-cred} and finds a $<$-complete assumption set (recall that this problem is non-trivial; $<$-complete assumption sets need not exist for a given framework; if none exist, the algorithm reports ``none exist''). Enumerating all $<$-complete assumptions sets can be achieved via Algorithm~\ref{alg:abap-com-enum}. 

\begin{algorithm}[h]
\caption{Finding a $<$-complete assumption set}
\label{alg:abap-com-find}
\begin{algorithmic}[1]
    \REQUIRE \abaplus{} framework $F=(\mathcal{L},\mathcal{R},\mathcal{A},\contraryempty, \leq)$
    \ENSURE return a $<$-complete assumption set of $F$ if one exists, Unsatisfiable otherwise
    \STATE{$\pi_{\mathit{cand}} := \mathtt{ABA^+}(F) \cup \pi_{cf}\cup \aspmodule{\mathit{undefeated}}^+ \cup \aspmodule{prune}^+$}
    \STATE{$\pi_{\mathit{check1}} := \mathtt{ABA^+}(F) \cup \aspmodule{defended}^+ \cup \aspmodule{suspect-defeat}^+$}
    \STATE{$\pi_{\mathit{check2}} := \mathtt{ABA^+}(F) \cup \aspmodule{com}^+ \cup \aspmodule{suspect-defeat}^+$}
    \STATE{$\algorithmicwhile\ \pi_{\mathit{cand}}$ is satisfiable $\algorithmicdo$}\label{alg:abap-com-find-2}
    \STATE{\hspace{\algorithmicindent} Let $I$ be the found answer set; $\mathit{flag}:=true$}
    \STATE{\hspace{\algorithmicindent} $\algorithmicif\ \pi_{\mathit{check1}} \cup {\bf undefeated}(I) \cup \aspin(I)$ unsat. $\algorithmicthen$}
    \STATE{\hspace{\algorithmicindent}\hspace{\algorithmicindent}$\algorithmicfor$ each $u\in \mathcal{A}$ such that ${\bf undefeated}(a)\in I\ \algorithmicdo$}
\STATE{\hspace{\algorithmicindent}\hspace{\algorithmicindent}\hspace{\algorithmicindent} $\algorithmicif\ \pi_{\mathit{check2}}
    \cup {\bf undefeated}(I)
    \cup \{{\bf target}(u)\}
    \cup {\bf in}(I)$ is\\\hspace{\algorithmicindent}\hspace{\algorithmicindent}\hspace{\algorithmicindent} unsatisfiable $\algorithmicthen\ \mathit{flag}:=false$; break} 
    \STATE{\hspace{\algorithmicindent}\hspace{\algorithmicindent}$\algorithmicif\ \mathit{flag}=true\ \algorithmicthen\ \algorithmicreturn\ I$}
    \STATE{\hspace{\algorithmicindent} $\pi_{\mathit{cand}} := \pi_{\mathit{cand}} \cup \{\mathit{constr}(\aspout(I) \cup \aspin(I))\}$}
    \RETURN{\emph{none exist}}
\end{algorithmic}
\end{algorithm}

\begin{algorithm}[!t]
\caption{Assumption set enumeration under $<$-complete semantics}
\label{alg:abap-com-enum}
\begin{algorithmic}[1]
    \REQUIRE \abaplus{} framework $F=(\mathcal{L},\mathcal{R},\mathcal{A},\contraryempty, \leq)$
    \ENSURE return all $<$-complete assumption sets of $F$
    \STATE{$\pi_{\mathit{cand}} := \mathtt{ABA^+}(F) \cup \pi_{cf}\cup \aspmodule{\mathit{undefeated}}^+ \cup \aspmodule{prune}^+$}
    \STATE{$\pi_{\mathit{check1}} := \mathtt{ABA^+}(F) \cup \aspmodule{defended}^+ \cup \aspmodule{suspect-defeat}^+$}
    \STATE{$\pi_{\mathit{check2}} := \mathtt{ABA^+}(F) \cup \aspmodule{com}^+ \cup \aspmodule{suspect-defeat}^+$}
    \STATE{$\algorithmicwhile\ \pi_{\mathit{cand}}$ is satisfiable $\algorithmicdo$}
    \STATE{\hspace{\algorithmicindent} Let $I$ be the found answer set; $\mathit{flag}:=true$}
    \STATE{\hspace{\algorithmicindent} $\algorithmicif\ \pi_{\mathit{check1}} \cup {\bf undefeated}(I) \cup \aspin(I)$ unsat. $\algorithmicthen$}
    \STATE{\hspace{\algorithmicindent}\hspace{\algorithmicindent}$\algorithmicfor$ each $u\in \mathcal{A}$ such that ${\bf undefeated}(a)\in I\ \algorithmicdo$}    \STATE{\hspace{\algorithmicindent}\hspace{\algorithmicindent}\hspace{\algorithmicindent} $\algorithmicif\ \pi_{\mathit{check2}}
    \cup {\bf undefeated}(I)
    \cup \{{\bf target}(u)\}
    \cup {\bf in}(I)$
  is\\\hspace{\algorithmicindent}\hspace{\algorithmicindent}\hspace{\algorithmicindent} unsatisfiable $\algorithmicthen\ \mathit{flag}:=false$; break}
    \STATE{\hspace{\algorithmicindent}\hspace{\algorithmicindent}$\algorithmicif\ \mathit{flag}=true\ \algorithmicthen\ E := E\cup\{I\}$}
    \STATE{\hspace{\algorithmicindent} $\pi_{\mathit{cand}} := \pi_{\mathit{cand}} \cup \{\mathit{constr}(\aspout(I) \cup \aspin(I))\}$}
    \RETURN{$E$}
\end{algorithmic}
\end{algorithm}
\newpage

\section{Experimental Results}
\label{appendix:exp}

\begin{figure}[!b]
    \centering
    \includegraphics[width=.63\linewidth]{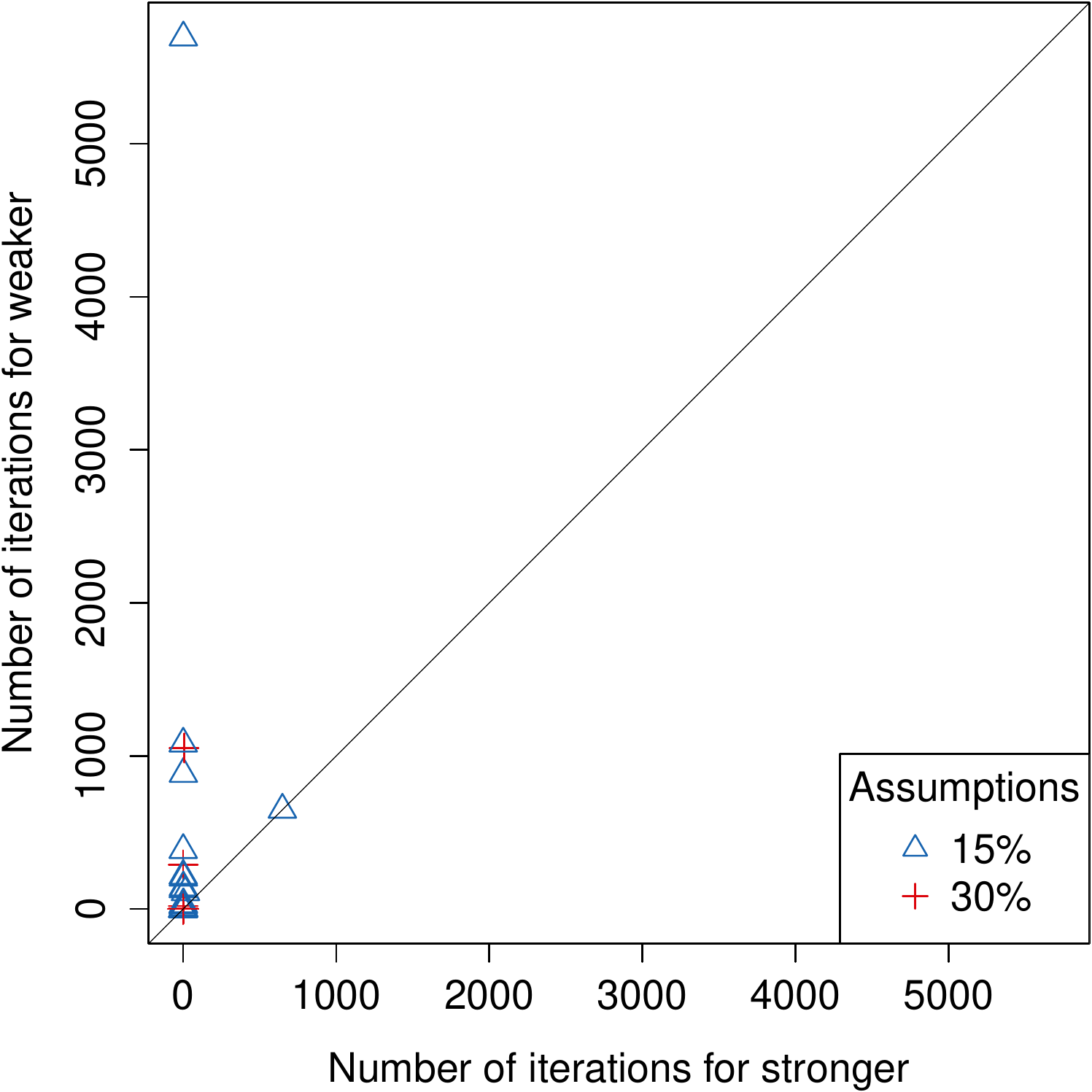}
    \caption{Comparison of number of iterations using the weaker and stronger abstraction under $<$-$\comp$ on the task of credulous reasoning.}
\label{fig:com-cred-iter}
\end{figure}

Figure~\ref{fig:com-cred-iter} shows the number of candidates found for instances which each version solved when considering the task of credulous acceptance.
Similarly to finding one assumption set (Figure~\ref{fig:com-extra} left), using the stronger refinement for the CEGAR algorithm for $<$-complete semantics drastically reduces the number of iterations needed to find the solution.
This supports the conclusion that reducing the number of candidates is at least a partial reason for the improvement in solving efficiency when using the stronger abstraction for $<$-complete semantics.

\end{document}